\ifdefined\pdftexversion\pdfoutput=1\fi
\documentclass[10pt,twocolumn,letterpaper]{article}
\usepackage[pagenumbers]{paperstyle}
\usepackage{amsmath,amssymb,booktabs,graphicx,multirow}
\definecolor{linkblue}{rgb}{0.16,0.37,0.55}
\usepackage[pagebackref,breaklinks,colorlinks,allcolors=linkblue]{hyperref}
\newcommand{\epm}{EPM}

\newcommand{\sg}{\operatorname{sg}}
\title{Perception into Imagination:\\ Executable Memory for World Models}
\author{Xianyao Li, Ruitong Tian, Rui Min, Fang Xu, Jing Du}
\begin{document}
\maketitle
\begin{abstract}
Physical agents must use what they previously saw or touched after visibility or contact is lost. Their actions change the implications of these remembered facts. We introduce executable perceptual memory (\epm), a world-model framework that advances explicit perceptual state alongside neural latent dynamics during imagination. Its evolving read-out informs reward prediction, value estimation and policy learning. We align reward inputs with the transitions that produced them and supervise the reward head on open-loop rollouts of recorded actions. On Memory Maze $13\times13$, EPM reaches 30.37 return at 1M interactions, compared with 23.37 when the same evolving read-out informs only policy and value on a common perceptual front end: a gain of 7.00 [3.80, 10.21]. The complete method also improves control on $9\times9$ and in MiniWorld. Probes of remembered, out-of-view arrivals show stronger reward anticipation in both visual implementations. Concentration, tactile recovery and real tactile frames demonstrate complementary roles of remembered facts in imagined-policy training, motion-dependent decisions and sensory-content retention. EPM connects persistent perception to the predicted consequences and learning signals of a world model for Physical AI.
\end{abstract}

\section{Introduction}
\label{sec:intro}
A robot can lose sight of a landmark when it turns, or lose tactile contact with an object when it withdraws. What it previously perceived remains useful, but the implications change with its actions: the landmark acquires a new bearing, and the remembered contact a new relative position. For Physical AI, using perception therefore requires both preserving facts and reasoning about their consequences in possible futures.

World models provide a setting for this reasoning: they predict action consequences and train behaviors on imagined trajectories~\cite{ha2018worldmodels,hafner2019planet,hafner2020dreamer,hafner2023dreamerv3,hansen2024tdmpc2}. Recurrent states encode history, while spatial memories and maps maintain explicit structure~\cite{parisotto2018neuralmap,chaplot2020activeslam,li2024memonav}. A further question is how maintained facts enter the futures used for learning: their relations to the agent change along a rollout, and these changes must inform predicted rewards and values as well as actions.

\begin{figure*}[t]
\centering\includegraphics[width=\textwidth]{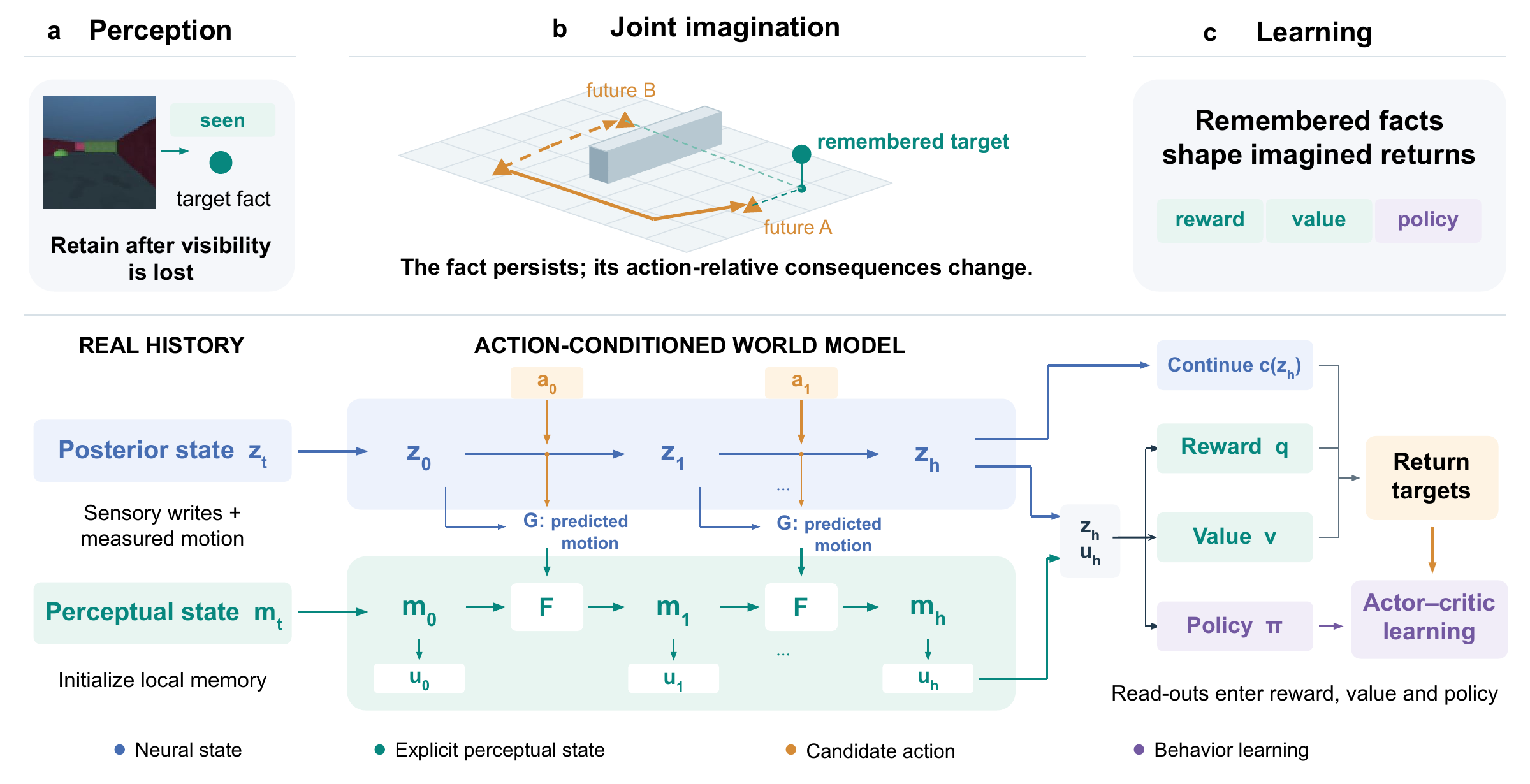}
\caption{\textbf{Executable perceptual memory within a world model.} Perception maintains facts beyond immediate visibility. Policy actions advance neural dynamics and explicit local state in parallel, coupled by predicted motion. The evolving read-out informs reward prediction and actor--critic learning; rewards, continuation and values form imagined return targets. Reward training uses transition-aligned inputs and open-loop supervision (Sec.~\ref{sec:method}). Above: an example observation and schematic futures for a remembered landmark. Below: the joint rollout and its evolving read-out $u_h$.}
\label{fig:overview}
\end{figure*}

We introduce \emph{executable perceptual memory} (\epm). A perceptual writer maintains explicit state during real interaction. At an imagination anchor, a local state is initialized from that history and advanced using candidate actions and predicted motion. Its read-out complements the neural latent state in reward prediction, value estimation and policy learning (Fig.~\ref{fig:overview}). A remembered fact can thus affect both the action under consideration and the return target used to learn that action.

The learning interface connects this joint state to the signals that train behavior. In our DreamerV3 implementation, a transition-aligned read-out refers to the fact relevant to each reward, before the next observation writes new evidence or changes the goal. An additional open-loop loss trains the reward head on prior rollouts of recorded actions. Together, these inputs and objectives make remembered relations available when learning rewards from real transitions and predicting them in imagination.

EPM improves control over a policy/value-only memory interface on a common front end in Memory Maze $9\times9$ and $13\times13$, and in MiniWorld. At $13\times13$, it reaches 30.37 return at 1M interactions, compared with 23.37 for that interface and 14.18 for observation-only memory access. Controlled probes in both visual implementations show stronger reward anticipation at remembered, out-of-view arrivals. Comparisons with DreamerV3 and R2I establish the approach's competitive context. Concentration, tactile recovery and real tactile frames test the same perception--memory--decision principle through complementary tasks.

We make three contributions. First, we formulate executable perceptual memory: explicit perceptual state that candidate actions advance alongside neural world-model dynamics, with its read-outs available throughout imagination. Second, we connect these read-outs to reward and actor--critic learning through transition-aligned reward inputs and open-loop supervision. Third, matched-information comparisons establish control gains and conditional component effects in two visual implementations, while symbolic and tactile tasks test complementary uses of persistent facts.

\section{Related work}
\label{sec:related}
\paragraph{World models and learning through imagination.}
World Models~\cite{ha2018worldmodels}, PlaNet~\cite{hafner2019planet} and Dreamer~\cite{hafner2020dreamer,hafner2021dreamerv2,hafner2023dreamerv3} learn predictive representations for planning or imagined behavior learning. Value Prediction Networks~\cite{oh2017vpn}, MuZero~\cite{schrittwieser2020muzero} and TD-MPC2~\cite{hansen2024tdmpc2} emphasize reward and value predictions for control. Dreamer 4~\cite{hafner2025dreamer4}, iVideoGPT~\cite{wu2024ivideogpt}, Navigation World Models~\cite{bar2025nwm}, V-JEPA 2~\cite{assran2025vjepa2} and Cosmos~\cite{nvidia2025cosmos} extend predictive modeling toward video and physical interaction. Multi-step objectives shape latent dynamics and belief states~\cite{hafner2019planet,gregor2019shaping}; DreamSmooth studies reward prediction under sparse rewards~\cite{lee2024dreamsmooth}. EPM connects persistent perceptual facts to the predicted rewards and values used for learning.

\paragraph{Memory under partial observability.}
Transformer policies~\cite{parisotto2020gtrxl}, structured sequence models~\cite{gu2022s4}, S4WM~\cite{deng2023s4wm} and R2I~\cite{samsami2024r2i} strengthen neural memory. Long-context video models preserve past observations~\cite{po2025longcontext,xiao2025worldmem,wu2025spatialmemory}; concurrent TT-MBRL encodes episode memory in model parameters at test time~\cite{burchi2026ttmbrl}. DIACRITIC studies recurrent memory sufficient to reproduce expert behavior~\cite{diacritic2026}. Cognitive Mapping and Planning~\cite{gupta2017cmp}, Neural Map~\cite{parisotto2018neuralmap}, Active Neural SLAM~\cite{chaplot2020activeslam} and MemoNav~\cite{li2024memonav} organize spatial history; Memoir retrieves history using imagined futures~\cite{xu2026memoir}. GTM-SM couples spatial memory to agent dynamics~\cite{fraccaro2018gtmsm}, and FloWM transforms latent memory with self-motion and object motion~\cite{lillemark2026flowm}. Concurrent ScanDreamer advances pose over a frozen map and analytically computes selected reward and termination signals~\cite{gao2026scandreamer}. EPM instead supplies executed fact read-outs to a learned reward model, using transition-aligned and open-loop training. Our comparisons hold the memory observation common and isolate how its read-outs enter imagination and learning.

\paragraph{Structured dynamics and grounded prediction.}
Interaction Networks~\cite{battaglia2016interaction}, SE3-Nets~\cite{byravan2017se3} and structured world models~\cite{kipf2020cswm} exploit object and geometric structure. Code world models~\cite{codeworldmodel} and Programmable World Model~\cite{programmableworldmodel} express transitions as executable structure. SearchWorld grounds search in spatial values~\cite{ji2026searchworld}; DreamLedger uses execution outcomes to assess reliance on model predictions~\cite{li2026dreamledger}. EPM specifies how perceptually maintained facts are executed and read by reward and behavior-learning interfaces.

\paragraph{Touch as persistent perceptual evidence.}
Optical tactile sensors recover local geometry and contact information~\cite{yuan2017gelsight}. Tactile Gym~\cite{lin2022tactilegym}, Taxim~\cite{si2022taxim} and TacSL~\cite{akinola2024tacsl} support simulation and learning, while Tactile MNIST~\cite{tactilemnist2025} supplies real observations for active perception. Visuo-tactile world models improve object permanence in predicted rollouts~\cite{higuera2026vtwm}. We study how tactile information persists beyond contact: explicit state advances along candidate motions, and compact sensory memory preserves content for later queries.

\section{Executable perceptual memory}
\label{sec:method}
\subsection{Persistent facts and action-dependent relations}
Consider a partially observed process with observations $o_t$, actions $a_t$, rewards $y_t$ and an optional goal $g_t$. We represent history by a neural state $z_t$ and explicit perceptual state $m_t$. The latter stores quantities with defined persistence and update rules: landmark estimates, object-relative poses or observed card identities. At interaction time, their values are estimated from the agent's observation stream. Privileged simulator state is not supplied to the policy.

Real interaction combines state propagation with a perceptual write:
\begin{equation}
 m_t=W\!\left(C(m_{t-1},\delta_t),E(o_t)\right),\qquad
 u_t=R(m_t,g_t).
 \label{eq:write}
\end{equation}
Here $E$ extracts sensory evidence, $\delta_t$ is measured motion, $C$ carries the stored state, and $W$ incorporates new evidence. The query $R$ gives a task-relevant read-out $u_t$. A landmark may remain fixed in world coordinates while its bearing changes with the agent's pose. This distinction lets the model preserve a fact while executing its consequences under movement.

\subsection{Joint imagination}
At a real anchor $t$, the posterior initializes $\tilde z_0$, and a projection of history initializes local state $\tilde m_0=P(m_t,g_t)$. The policy acts on their joint read-out. Each imagined step advances both components:
\begin{equation}
\begin{aligned}
 \tilde a_h&\sim\pi_\psi(\cdot\mid\tilde z_h,u_h),\\
 \tilde z_{h+1}&\sim p_\theta(\cdot\mid\tilde z_h,\tilde a_h),\\
 \widehat\delta_h&=G_\theta(\tilde z_h,\tilde a_h),\\
 \tilde m_{h+1}&=F(\tilde m_h,\widehat\delta_h,\tilde a_h),\quad
 u_{h+1}=R(\tilde m_{h+1},g_t).
\end{aligned}
\label{eq:imagine}
\end{equation}
The neural model learns dynamics from experience; the executable update applies candidate actions to selected history. The paths interact through policy actions and predicted motion. The neural transition is conditioned on the latent state and action.

In navigation, $P$ extracts the current target's relative geometry and age. Predicted translation and rotation advance this local relation; new observations and target switches are outside that short-rollout update. A shared motion head learns the subsequent odometry measurement from the pre-action neural feature and action with a squared-error objective. Motion predictions used by $F$ are stopped-gradient. In the tactile instance, the carried quantity is object-relative pose; in Concentration, known identities constrain a simulator of the public rules. The framework specifies where executable structure enters, while each instance defines the structure it can propagate.

\subsection{From imagined facts to learning signals}
The navigation method supplies $[\tilde z_h,u_h]$ to both the reward head $q_\omega$ and actor--critic. Continuation $c_\theta$ reads neural features alone:
\begin{equation}
 \hat y_h=q_\omega(\tilde z_h,u_h),\quad
 \hat c_h=c_\theta(\tilde z_h),\quad
 \hat v_h=v_\xi(\tilde z_h,u_h).
 \label{eq:targets}
\end{equation}
Predicted rewards, continuation and bootstrapped values form the usual Dreamer returns. The explicit state therefore informs the learning target as well as the behavior being optimized. Policy, critic, slow critic and replay value computation all retain direct read-out access.

\paragraph{Aligning facts with rewards.}
Reward $y_{t+1}$ is produced by action $a_t$. Its replay-side read-out must refer to that transition, before the new observation writes evidence or switches the target. We obtain it by advancing the preceding memory with measured motion:
\begin{equation}
 \bar u_{t+1}=R\!\left(F(P(m_t,g_t),\delta_{t+1},a_t),g_t\right).
 \label{eq:alignment}
\end{equation}
The posterior reward loss pairs $(z_{t+1},\bar u_{t+1})$ with $y_{t+1}$. At episode starts and at the first replay-sequence position, the additional slots are zeroed. In imagination, the head instead uses the read-out advanced with predicted motion along the imagined trajectory.

\paragraph{Training rewards on open-loop futures.}
From replay posterior anchors $t\in\mathcal S$, we roll the prior along the recorded actions for $H$ steps without further observations, and propagate the local memory in the same way. Let $b_{t,h}$ indicate that no episode reset has occurred between the anchor and step $h$. The added loss is
\begin{equation}
 \mathcal L_{\rm OL}=\frac{1}{|\mathcal S|}\sum_{t\in\mathcal S}
 \frac{\sum_{h=1}^{H}b_{t,h}\,
 \ell_\omega(y_{t+h};\sg[\tilde z_h,u_h])}
 {\max(1,\sum_{h=1}^{H}b_{t,h})}.
 \label{eq:ol}
\end{equation}
Here $\ell_\omega$ is the backbone reward-prediction loss. We use $H=15$, anchors every four replay steps and weight 1. The stop-gradient boundary makes this term train only the reward head, leaving representation, dynamics and motion learning under their existing objectives. The posterior reward loss is retained; Appendix~\ref{app:method} illustrates both training paths.

\paragraph{Perceptual implementation.}
A landmark detector and visual odometry maintain sparse target positions and agent pose. The read-out contains a known flag, sine/cosine bearing, log-distance and log-age. These perception networks are trained on separate calibration trajectories with simulator labels, then frozen; deployment uses images and actions. Memory Maze adds image-based heading correction, and MiniWorld models lateral motion. Every arm encodes and reconstructs the same memory observation and shares the same front end within an experiment. Appendix~\ref{app:method} details calibration, state updates and storage.

\section{Experimental setup}
\label{sec:setup}
\paragraph{Tasks and training.}
The principal learned-control tests use multi-goal Memory Maze~\cite{pasukonis2022memorymaze} at $9\times9$ and $13\times13$, and a MiniWorld~\cite{chevalier2023miniworld} $9\times9$ implementation with different rendering and kinematics. The DreamerV3-based agent uses the 12M-parameter configuration, 16 environments and a 2M-transition replay capacity. All policies train from scratch. Memory Maze endpoints are 1M interactions, with twelve initialization labels at $9\times9$ and six at $13\times13$; MiniWorld uses twelve initialization labels and a 2M endpoint. Each policy is evaluated on 48 common layout/target/spawn identities. Episode budgets are 1,000 actions for both $9\times9$ implementations and 3,000 for $13\times13$. Concentration, tactile simulation and real tactile frames test complementary stages of the framework (Sec.~\ref{sec:touch}).

\paragraph{Six integration choices.}
Table~\ref{tab:interfaces} names arms by their components. ``Held'' and ``evolved'' refer to direct read-outs during imagination. All arms have the same memory information in observation encoding, matched head dimensions and common initial parameters within a label. Zero-filled direct slots preserve parameter counts. The six-arm design is used at $9\times9$ and in MiniWorld; the $13\times13$ experiment trains the four non-held arms. ``Evolved'' supplies the evolving read-out to the policy and value only, without reward access or the open-loop loss.
\begin{table}[t]\centering\small
\caption{\textbf{Navigation arms.} All encode the same perceptual memory. Entries specify direct imagined read-outs. R = reward access; OL = open-loop reward loss. Reward access is tested with OL.}
\label{tab:interfaces}
\setlength{\tabcolsep}{4pt}
\begin{tabular}{lccc}\toprule
Arm & Policy/value & Reward & OL\\\midrule
Observation+OL & -- & -- & Yes\\
Held+OL & Held & -- & Yes\\
Held+R+OL & Held & Held & Yes\\
Evolved & Evolved & -- & No\\
Evolved+OL & Evolved & -- & Yes\\
\textbf{EPM} & Evolved & Evolved & Yes\\\bottomrule
\end{tabular}\end{table}

\paragraph{Comparisons and inference.}
The full-method contrast is EPM minus Evolved; reward access is EPM minus Evolved+OL; OL alone is Evolved+OL minus Evolved. The held arms measure reward access under holding. Return differences use paired two-sided 95\% $t$ intervals over initialization labels, with positive-label counts and sign-flip sensitivity analyses in the appendix. Each task is inferred separately; secondary contrasts are not multiplicity-adjusted. Budgets and primary endpoints are fixed before evaluation. Interaction budgets count policy training and exclude the shared offline perception calibration.

\paragraph{Reward-anticipation probes.}
Every arm is tested on common histories from EPM and Evolved+OL policies, with equal source-policy weight. We compare visible targets (VIS) with remembered, out-of-view targets whose arrival is anticipated by explicit state (MEM and RA). RA is evaluated per model, so the retained event subsets can differ. Net reward anticipation $A$ subtracts predicted mass in non-arrival windows from mass around first arrivals, weighting eligible horizon bins equally. It is not an accuracy or probability. The memory deficit is $D=A_{\rm VIS}-A_{\rm MEM\cap RA}$; Appendix~\ref{app:mechanism} gives the full definitions.

\section{Perceptual history improves learning and control}
\label{sec:vision}
\subsection{The complete method improves control}
\begin{figure*}[t]
\centering\includegraphics[width=\textwidth]{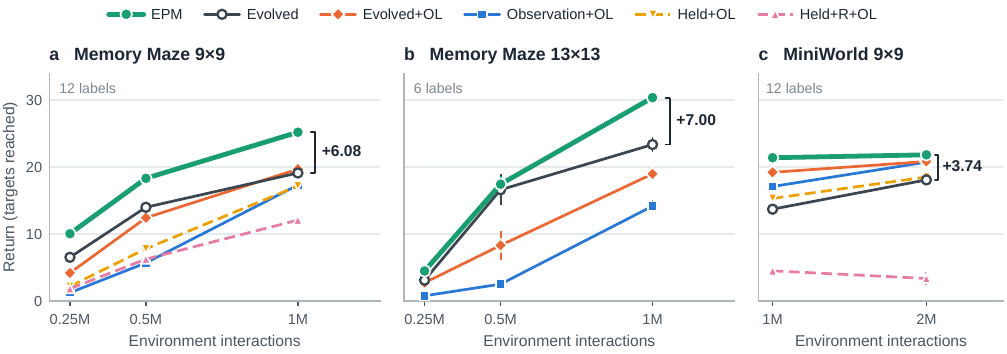}
\caption{\textbf{Control with executable perceptual memory.} Mean return at evaluation checkpoints; bars show one standard error over initialization labels. Each panel is a separate experiment with a common front end across its arms. Brackets mark the paired endpoint gain of EPM over Evolved, which lacks reward access and the open-loop loss. Table~\ref{tab:main_contrasts} gives endpoint returns and component effects.}
\label{fig:navigation}
\end{figure*}
On the larger $13\times13$ task, EPM reaches 30.37 return at 1M interactions, compared with 23.37 for the policy/value-only interface (Evolved) on the same front end. The gain is $7.00$ [$3.80$, $10.21$], positive on all six initialization labels. Against observation-only memory access, the gain is $16.19$ [$12.67$, $19.72$]: the same perceptual information is more useful when it participates directly in imagined prediction and learning. The $9\times9$ experiment independently gives $6.08$ [$4.46$, $7.71$] over Evolved, positive on all twelve labels.

The second visual implementation tests the learning design with different rendering and planar motion. In MiniWorld at 2M, EPM reaches 21.83 versus 18.09 for Evolved: $+3.74$ [$2.51$, $4.97$], positive on all twelve labels. The complete method also exceeds observation-only access in each experiment. Figure~\ref{fig:navigation} shows learning progress, and Table~\ref{tab:main_contrasts} gives endpoint returns and component effects.

\begin{table*}[t]\centering\small
\caption{\textbf{Control results and component effects.} Upper: mean return at the endpoint. Lower: pre-specified paired differences $\Delta$ with 95\% $t$ intervals over initialization labels; $n_{+}$ is the number of labels with a positive difference. Each task is a separate experiment with a common front end across its arms. R denotes direct reward access and OL the open-loop reward loss. The $13\times13$ experiment trains the four non-held arms. The interaction is the reward-access effect for evolved minus held read-outs.}
\label{tab:main_contrasts}
\setlength{\tabcolsep}{0pt}
\begin{tabular}{@{}l@{\hskip 10pt}r@{\hskip 4pt}l@{\hskip 6pt}c@{\hskip 14pt}r@{\hskip 4pt}l@{\hskip 6pt}c@{\hskip 14pt}r@{\hskip 4pt}l@{\hskip 6pt}c@{}}\toprule
 & \multicolumn{3}{c}{Memory Maze $9\times9$} & \multicolumn{3}{c}{Memory Maze $13\times13$} & \multicolumn{3}{c}{MiniWorld $9\times9$} \\
 & \multicolumn{3}{c}{1M, 12 labels} & \multicolumn{3}{c}{1M, 6 labels} & \multicolumn{3}{c}{2M, 12 labels} \\
\midrule
Observation+OL & \multicolumn{3}{c}{17.32} & \multicolumn{3}{c}{14.18} & \multicolumn{3}{c}{20.74} \\
Held+OL & \multicolumn{3}{c}{17.18} & \multicolumn{3}{c}{--} & \multicolumn{3}{c}{18.52} \\
Held+R+OL & \multicolumn{3}{c}{12.08} & \multicolumn{3}{c}{--} & \multicolumn{3}{c}{3.37} \\
Evolved & \multicolumn{3}{c}{19.12} & \multicolumn{3}{c}{23.37} & \multicolumn{3}{c}{18.09} \\
Evolved+OL & \multicolumn{3}{c}{19.69} & \multicolumn{3}{c}{18.98} & \multicolumn{3}{c}{20.83} \\
\textbf{EPM} & \multicolumn{3}{c}{\textbf{25.20}} & \multicolumn{3}{c}{\textbf{30.37}} & \multicolumn{3}{c}{\textbf{21.83}} \\
\midrule
Paired difference & \multicolumn{1}{c}{$\Delta$} & \multicolumn{1}{c}{95\% CI} & $n_{+}$ & \multicolumn{1}{c}{$\Delta$} & \multicolumn{1}{c}{95\% CI} & $n_{+}$ & \multicolumn{1}{c}{$\Delta$} & \multicolumn{1}{c}{95\% CI} & $n_{+}$ \\
\midrule
EPM $-$ Evolved & $+6.08$ & [$4.46$, $7.71$] & 12/12 & $+7.00$ & [$3.80$, $10.21$] & 6/6 & $+3.74$ & [$2.51$, $4.97$] & 12/12 \\
EPM $-$ Evolved+OL & $+5.51$ & [$4.46$, $6.57$] & 12/12 & $+11.39$ & [$7.89$, $14.89$] & 6/6 & $+1.00$ & [$-0.04$, $2.03$] & 10/12 \\
EPM $-$ Observation+OL & $+7.89$ & [$6.17$, $9.60$] & 12/12 & $+16.19$ & [$12.67$, $19.72$] & 6/6 & $+1.09$ & [$0.02$, $2.17$] & 9/12 \\
Evolved+OL $-$ Evolved & $+0.57$ & [$-1.55$, $2.70$] & 7/12 & $-4.39$ & [$-8.16$, $-0.61$] & 0/6 & $+2.74$ & [$1.16$, $4.33$] & 11/12 \\
Evolved+OL $-$ Held+OL & $+2.51$ & [$1.36$, $3.66$] & 11/12 & \multicolumn{3}{c}{--} & $+2.31$ & [$1.10$, $3.52$] & 11/12 \\
Held+R+OL $-$ Held+OL & $-5.10$ & [$-6.34$, $-3.86$] & 0/12 & \multicolumn{3}{c}{--} & $-15.15$ & [$-17.95$, $-12.36$] & 0/12 \\
Reward access $\times$ evolution & $+10.61$ & [$9.35$, $11.86$] & 12/12 & \multicolumn{3}{c}{--} & $+16.15$ & [$13.05$, $19.25$] & 12/12 \\
\bottomrule\end{tabular}
\end{table*}

\begin{figure*}[t]\centering
\includegraphics[width=\textwidth]{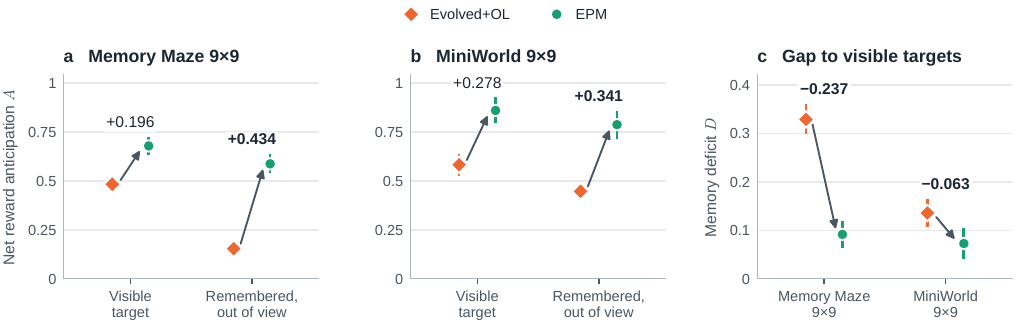}
\caption{\textbf{Reward anticipation from perceptual history.} (a,b) Net anticipation for visible targets and remembered, out-of-view targets satisfying each model's arrival-anticipation condition (RA). (c) Their difference, the memory deficit. Points and bars are means and 95\% $t$ intervals over twelve labels per implementation. Both arms have evolved policy/value read-outs and OL training; EPM additionally supplies reward read-outs. Numbers give paired differences.}
\label{fig:mechanism}\end{figure*}

\subsection{What do the components contribute?}
\paragraph{Reward access.}
Under common open-loop training, EPM exceeds Evolved+OL by $5.51$ [$4.46$, $6.57$] on $9\times9$ and $11.39$ [$7.89$, $14.89$] on $13\times13$, positive on every label. In MiniWorld, this increment is $1.00$ [$-0.04$, $2.03$] at 2M and remains unresolved. Thus the complete method improves control across all three tasks, with the reward-access contribution established in both Memory Maze sizes.

\paragraph{Action-dependent evolution.}
With common OL and no direct reward access, evolution improves return over holding by $2.51$ [$1.36$, $3.66$] on $9\times9$ and $2.31$ [$1.10$, $3.52$] in MiniWorld. Reward access under holding lowers return by $5.10$ [$3.86$, $6.34$] and $15.15$ [$12.36$, $17.95$], respectively; Held+R+OL reaches 3.37 in MiniWorld. Both reward-access arms use transition-aligned replay inputs; holding changes the rollout-side read-out. Access-by-evolution interactions are 10.61 and 16.15; the latter mainly reflects the held arm's degradation. These controls distinguish executing remembered relations from repeatedly supplying an unchanged read-out.

\paragraph{Open-loop supervision.}
Without direct reward access, OL gives $+0.57$ [$-1.55$, $2.70$] on $9\times9$, $-4.39$ [$-8.16$, $-0.61$] on $13\times13$, and $+2.74$ [$1.16$, $4.33$] in MiniWorld. The large $13\times13$ reward-access gain occurs where OL alone lowers return. Appendix~\ref{app:formal} provides full contrast tables.

\subsection{Remembered facts inform predicted rewards}
For Memory Maze $9\times9$, reward access raises remembered-arrival anticipation from 0.154 to 0.587. The paired increment is $0.434$ [$0.376$, $0.491$], positive on 12/12 labels. Visible anticipation also increases, from 0.482 to 0.679; the memory deficit decreases from 0.329 to 0.092, a paired reduction of 0.237 [0.211, 0.264]. The larger increase on remembered arrivals narrows the deficit beyond the overall rise in anticipation.

In MiniWorld, remembered-arrival anticipation rises from 0.447 to 0.787: $+0.341$ [$0.283$, $0.398$], again positive on 12/12 labels. Visible anticipation rises from 0.582 to 0.860, and the deficit falls by 0.063 [0.024, 0.102]. Across both visual implementations, the interface increases predicted reward at remembered arrivals and narrows the gap to visible arrivals, consistent with remembered facts reaching reward prediction.

\subsection{Comparison with DreamerV3 and R2I}
We compare with native DreamerV3 and R2I, and with R2I supplied with the same perception stream and maintained state (Table~\ref{tab:baselines}). A six-label experiment uses visual odometry without heading correction and a direct policy/value interface. It reaches 11.17 at 1M, versus 5.93 with memory accessed only through observation encoding: $+5.24$ [$1.05$, $9.43$]. Appendix~\ref{app:fresh} gives all checkpoints. Table~\ref{tab:baselines} separates this setting from the heading-corrected experiments.

This direct-interface agent has higher mean return than native DreamerV3 at 1M (2.90) and native R2I at 3M (6.74). R2I with the front end and maintained state reaches 9.36 at 3M; the training-seed difference from 11.17 is unresolved. With the shared heading-corrected front end, Evolved reaches 23.37 and EPM 30.37.

\begin{table}[t]\centering\small
\caption{\textbf{Comparison with external world-model agents on $13\times13$.} Mean returns with training budgets. External methods use three training seeds; our agents use six initialization labels. The upper group uses visual odometry without heading correction where a front end is supplied; the lower pair shares the heading-corrected front end and is a separate experiment.}
\label{tab:baselines}
\setlength{\tabcolsep}{5pt}
\begin{tabular}{@{}lrr@{}}\toprule
Method & Interactions & Return\\\midrule
Native DreamerV3 & 1M & 2.90\\
Native R2I & 3M & 6.74\\
R2I + front end & 3M & 6.08\\
R2I + front end + state & 3M & 9.36\\
Policy/value-only interface & 1M & 11.17\\\midrule
\multicolumn{3}{@{}l}{\textit{With heading correction}}\\
Policy/value-only interface (Evolved) & 1M & 23.37\\
\textbf{EPM} & 1M & \textbf{30.37}\\\bottomrule
\end{tabular}\end{table}

\section{Remembered facts and physical interaction}
\label{sec:touch}
The following instances test complementary stages: respecting facts during imagined training, carrying state through lost contact, and retaining real sensory content for later use.
\subsection{Remembered facts improve imagined-policy training}
\begin{figure*}[t]
\centering\includegraphics[width=\textwidth]{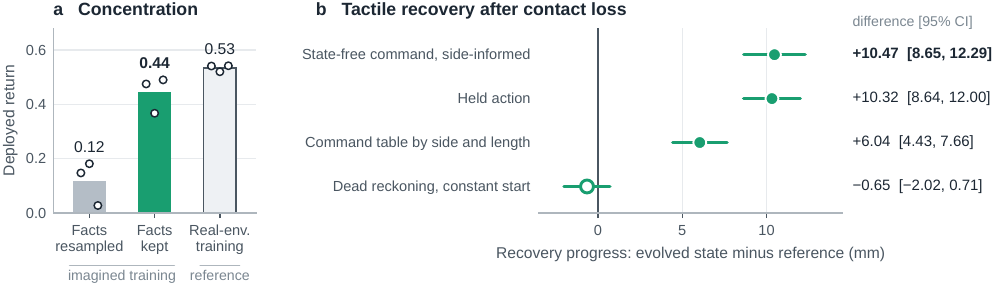}
\caption{\textbf{History matters at different stages of decision making.} (a) Deployed Concentration return of policies trained on imagined trajectories with remembered card identities resampled or kept, and of a policy trained in the real environment; dots are training seeds. (b) Tactile recovery progress of the evolved-state chooser minus each reference on 48 new worlds with varying excursions; bars show 95\% paired $t$ intervals over worlds. The primary comparison uses a state-free command informed of excursion side; the hollow marker is a reference that also carries state by motion.}
\label{fig:facts_touch}
\end{figure*}
ConcentrationHard~\cite{morad2023popgym} gives a discrete instance of the same question. A key--value memory stores observed card identities. During imagination, a simulator executes the public game rules while retaining those identities and sampling unknown cards. A matched control uses the same rules but resamples all identities. The policy is optimized entirely on imagined rollouts from public-observation prefixes, then deployed in the real game.

Retaining remembered facts yields deployed returns of 0.475, 0.367 and 0.490 over three training seeds, compared with 0.147, 0.181 and 0.027 when all identities are resampled. Real-environment training gives 0.541, 0.520 and 0.542 (256 evaluation episodes per policy). The supplied rule simulator differs from the neural navigation model, but the comparison isolates a shared methodological point: the futures used for training should respect the facts already acquired by the agent.

\subsection{Tactile history supports future branches and recovery}
In Tactile Gym 2.0~\cite{lin2022tactilegym}, an estimator writes object-relative state from tactile observations while contact is available. Kinematic updates carry this state after contact is lost and advance it along candidate actions. On twelve previously unexposed worlds, explicit evolution reduces pre-contact prediction error by 0.51 mm [0.28, 0.73] and raises agreement with the best executed branch from 0.02 to 0.64.

We next test recovery choices after a retreat and sideways excursion. A chooser evaluates 128 fixed open-loop paths using an evolved state and a fitted outcome model, then hands over to a feedback law at re-contact. On 48 new worlds with varying excursion lengths, it achieves 34.80 mm of progress per recovery window, versus 24.33 mm for a state-free command selected by excursion side. The primary difference is 10.47 [8.65, 12.29] mm, positive on 46/48 worlds. It also exceeds a held-action reference by 10.32 [8.64, 12.00] mm and a command table indexed by side and excursion length by 6.04 [4.43, 7.66] mm (Fig.~\ref{fig:facts_touch}).

A constant-start dead-reckoning reference reaches 35.46 mm; evolved minus this reference is $-0.65$ [$-2.02$, $0.71$] mm and remains unresolved. The results support motion-carried state for candidate-path selection; this comparison does not isolate tactile initialization. Appendix~\ref{app:tactile} gives protocols, splits and additional comparisons.

\subsection{Real tactile observations survive in compact memory}
\begin{figure}[t]
\centering\includegraphics[width=\linewidth]{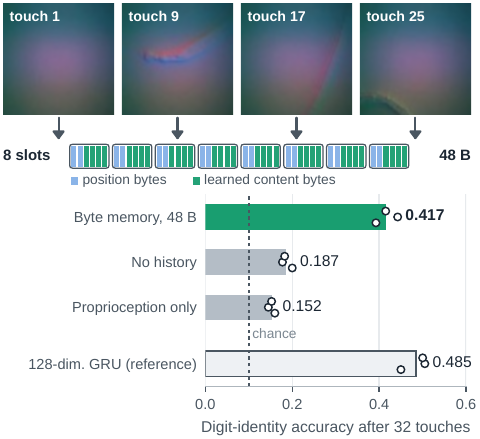}
\caption{\textbf{Real perceptual content retained for later queries.} Top: fixed-index, unmodified training-set GelSight Mini frames (Real Tactile MNIST, Schneider et al.; CC BY 2.0), written into eight six-byte slots. Bottom: digit-identity accuracy of a late query after 32 touches on held-out objects; dots are training seeds.}
\label{fig:real_touch}
\end{figure}

Real Tactile MNIST contains GelSight Mini images collected with a robot manipulator~\cite{tactilemnist2025}. We encode contact frames and their positions into eight slots with 48 bytes of payload plus up to 9 bytes of validity and age metadata. A late query asks for digit identity or material at a held-out position. Evaluation uses 100 held-out objects; background subtraction uses a per-round offline background computed from all 256 touches. Figure~\ref{fig:real_touch} shows original training-set frames and the held-out query results.

After 32 touches, the byte memory achieves mean identity accuracy 0.417, compared with 0.187 without history and 0.152 with proprioception and the shared image-derived write gate; individual-seed values appear in Fig.~\ref{fig:real_touch}. Removing content bytes returns accuracy to chance, and larger payloads improve retention. A 128-dimensional, non-quantized GRU reference reaches 0.485. For material queries, the byte memory exceeds this control on two of three seeds, with the third unresolved. Compact memory thus retains useful real sensory content.

\section{Discussion and conclusion}
\label{sec:discussion}
Executable perceptual memory gives world models a way to use facts that outlast their immediate observation. Perception maintains the facts, candidate actions advance their local consequences, and reward and behavior-learning interfaces use the resulting state. Control gains in two visual implementations and improved anticipation of remembered arrivals support this connection from perceptual history to imagined learning signals.

The component experiments identify where this connection matters. Action-dependent evolution improves on held read-outs, while open-loop supervision depends on the task and the information supplied to the reward head. The current local navigation state follows a remembered target during short rollouts. Extending it to anticipated observations, target changes and richer contact transitions is a natural direction for the framework.

The symbolic and tactile instances show complementary uses of the same principle: known facts constrain imagined-policy training, carried state supports recovery choices, and compact memory preserves real sensory content. These results motivate world models for Physical AI in which perceptual history remains available as actionable state throughout imagination and learning.

{\small\bibliographystyle{ieeenat_fullname}\bibliography{references}}

@article{hafner2023dreamerv3,
  title = {Mastering Diverse Control Tasks through World Models},
  author = {Hafner, Danijar and Pasukonis, Jurgis and Ba, Jimmy and Lillicrap, Timothy},
  journal = {Nature}, volume = {640}, pages = {647--653}, year = {2025},
  doi = {10.1038/s41586-025-08744-2}
}

@inproceedings{samsami2024r2i,
  title     = {Mastering Memory Tasks with World Models},
  author    = {Samsami, Mohammad Reza and Zholus, Artem and Rajendran, Janarthanan and Chandar, Sarath},
  booktitle = {Int. Conf. Learn. Represent.},
  year      = {2024},
  note      = {arXiv:2403.04253}
}

@inproceedings{pasukonis2022memorymaze,
  title     = {Evaluating Long-Term Memory in {3D} Mazes},
  author    = {Pasukonis, Jurgis and Lillicrap, Timothy and Hafner, Danijar},
  booktitle = {Int. Conf. Learn. Represent.},
  year      = {2023},
  note      = {arXiv:2210.13383}
}

@inproceedings{morad2023popgym,
  title     = {{POPGym}: Benchmarking Partially Observable Reinforcement Learning},
  author    = {Morad, Steven and Kortvelesy, Ryan and Bettini, Matteo and Liwicki, Stephan and Prorok, Amanda},
  booktitle = {Int. Conf. Learn. Represent.},
  year      = {2023},
  note      = {arXiv:2303.01859}
}

@inproceedings{deng2023s4wm,
  title     = {Facing Off World Model Backbones: {RNNs}, {Transformers}, and {S4}},
  author    = {Deng, Fei and Park, Junyeong and Ahn, Sungjin},
  booktitle = {Adv. Neural Inform. Process. Syst.},
  year      = {2023},
  note      = {arXiv:2307.02064}
}

@article{si2022taxim,
  title   = {Taxim: An Example-Based Simulation Model for {GelSight} Tactile Sensors},
  author  = {Si, Zilin and Yuan, Wenzhen},
  journal = {IEEE Robot. Autom. Lett.},
  volume  = {7}, number = {2}, pages = {2361--2368},
  year    = {2022},
  note    = {arXiv:2109.04027}
}

@article{akinola2024tacsl,
  title = {{TacSL}: A Library for Visuotactile Sensor Simulation and Learning},
  author = {Akinola, Iretiayo and Xu, Jie and Carius, Jan and Fox, Dieter and Narang, Yashraj},
  journal = {IEEE Transactions on Robotics}, volume = {41}, pages = {2645--2661}, year = {2025},
  doi = {10.1109/TRO.2025.3547267}
}

@inproceedings{tactilemnist2025,
  title = {Tactile {MNIST}: Benchmarking Active Tactile Perception},
  author = {Schneider, Tim and Duret, Guillaume and de Farias, Cristiana and Calandra, Roberto and Chen, Liming and Peters, Jan},
  booktitle = {Advances in Neural Information Processing Systems, Evaluations and Datasets Track},
  year = {2026}, note = {arXiv:2506.06361}
}

@article{diacritic2026,
  title   = {Minimal Recurrent Behavioral Memory for Imitation under Partial Observability},
  author  = {Li, Xianyao and Xu, Fang and Min, Rui and Tian, Ruitong and Du, Jing},
  journal = {arXiv preprint arXiv:2609.25757},
  year    = {2026},
}

@inproceedings{codeworldmodel,
  title     = {Generating Code World Models with Large Language Models Guided by {Monte Carlo} Tree Search},
  author    = {Dainese, Nicola and Merler, Matteo and Alakuijala, Minttu and Marttinen, Pekka},
  booktitle = {Adv. Neural Inform. Process. Syst.},
  year      = {2024},
  note      = {arXiv:2405.15383}
}

@article{programmableworldmodel,
  title   = {Programmable World Model},
  author  = {Huang, Zheng-Hui and Lin, Guixu and Lin, Jiacheng and Huang, Yi-Chuan and Yu, Ruihan and Niu, Muyao and Yang, Siqi and Liu, Yu-Lun and Chuang, Yung-Yu and Zhang, Kaipeng and Wang, Zhixiang},
  journal = {arXiv preprint arXiv:2609.10540},
  year    = {2026}
}

@inproceedings{gu2022s4,
  title     = {Efficiently Modeling Long Sequences with Structured State Spaces},
  author    = {Gu, Albert and Goel, Karan and R{\'e}, Christopher},
  booktitle = {Int. Conf. Learn. Represent.},
  year      = {2022}
}

@inproceedings{parisotto2020gtrxl,
  title     = {Stabilizing Transformers for Reinforcement Learning},
  author    = {Parisotto, Emilio and Song, Francis and Rae, Jack and Pascanu, Razvan and Gulcehre, Caglar and Jayakumar, Siddhant and Jaderberg, Max and Kaufman, Rapha{\"e}l Lopez and Clark, Aidan and Noury, Seb and Botvinick, Matthew and Heess, Nicolas and Hadsell, Raia},
  booktitle = {Int. Conf. Mach. Learn.},
  year      = {2020}
}

@inproceedings{gregor2019shaping,
  title     = {Shaping Belief States with Generative Environment Models for {RL}},
  author    = {Gregor, Karol and Jimenez Rezende, Danilo and Besse, Frederic and Wu, Yan and Merzic, Hamza and van den Oord, Aaron},
  booktitle = {Adv. Neural Inform. Process. Syst.},
  year      = {2019}
}

@article{ji2026searchworld,
  title   = {{SearchWorld}: Spatial Value-Grounded Imagination for {UAV} Object Search via World Models},
  author  = {Ji, Yatai and Zhu, Zhengqiu and Zhao, Yong and Hu, Yue and Yao, Fanglong and Gao, Chen and Zhu, Pengfei and Yin, Quanjun},
  journal = {arXiv preprint arXiv:2610.09335},
  year    = {2026}
}

@article{xu2026memoir,
  title   = {Dream to Recall: Imagination-Guided Experience Retrieval for Memory-Persistent Vision-and-Language Navigation},
  author  = {Xu, Yunzhe and Pan, Yiyuan and Liu, Zhe},
  journal = {IEEE Transactions on Pattern Analysis and Machine Intelligence},
  volume  = {48}, number = {8}, pages = {9035--9049},
  year    = {2026},
  doi     = {10.1109/TPAMI.2026.3679426}
}

@inproceedings{li2024memonav,
 title={{MemoNav}: Working Memory Model for Visual Navigation},
 author={Li, Hongxin and Wang, Zeyu and Yang, Xu and Yang, Yuran and Mei, Shuqi and Zhang, Zhaoxiang},
 booktitle={IEEE Conf. Comput. Vis. Pattern Recog.},
 pages={17913--17922}, year={2024}
}

@inproceedings{bar2025nwm,
 title={Navigation World Models},
 author={Bar, Amir and Zhou, Gaoyue and Tran, Danny and Darrell, Trevor and LeCun, Yann},
 booktitle={IEEE Conf. Comput. Vis. Pattern Recog.},
 pages={15791--15801},
 year={2025}
}

@inproceedings{hafner2020dreamer,
 title={Dream to Control: Learning Behaviors by Latent Imagination},
 author={Hafner, Danijar and Lillicrap, Timothy and Ba, Jimmy and Norouzi, Mohammad},
 booktitle={Int. Conf. Learn. Represent.}, year={2020}, note={arXiv:1912.01603}
}

@inproceedings{hansen2024tdmpc2,
 title={{TD-MPC2}: Scalable, Robust World Models for Continuous Control},
 author={Hansen, Nicklas and Su, Hao and Wang, Xiaolong},
 booktitle={Int. Conf. Learn. Represent.}, year={2024}, note={arXiv:2310.16828}
}

@inproceedings{wu2024ivideogpt,
 title={{iVideoGPT}: Interactive {VideoGPTs} are Scalable World Models},
 author={Wu, Jialong and Yin, Shaofeng and Feng, Ningya and He, Xu and Li, Dong and Hao, Jianye and Long, Mingsheng},
 booktitle={Adv. Neural Inform. Process. Syst.}, year={2024}, note={arXiv:2405.15223}
}

@article{lin2022tactilegym,
 title={{Tactile Gym} 2.0: Sim-to-Real Deep Reinforcement Learning for Comparing Low-Cost High-Resolution Robot Touch},
 author={Lin, Yijiong and Lloyd, John and Church, Alex and Lepora, Nathan F.},
 journal={IEEE Robot. Autom. Lett.}, volume={7}, number={4}, pages={10754--10761}, year={2022},
 doi={10.1109/LRA.2022.3195195}
}

@article{ha2018worldmodels,
 title={World Models}, author={Ha, David and Schmidhuber, J{\"u}rgen},
 journal={arXiv preprint arXiv:1803.10122}, year={2018}
}

@inproceedings{hafner2019planet,
 title={Learning Latent Dynamics for Planning from Pixels},
 author={Hafner, Danijar and Lillicrap, Timothy and Fischer, Ian and Villegas, Ruben and Ha, David and Lee, Honglak and Davidson, James},
 booktitle={Int. Conf. Mach. Learn.}, pages={2555--2565}, year={2019}
}

@inproceedings{hafner2021dreamerv2,
 title={Mastering {Atari} with Discrete World Models},
 author={Hafner, Danijar and Lillicrap, Timothy and Norouzi, Mohammad and Ba, Jimmy},
 booktitle={Int. Conf. Learn. Represent.}, year={2021}
}

@article{hafner2025dreamer4,
 title={Training Agents Inside of Scalable World Models},
 author={Hafner, Danijar and Yan, Wilson and Lillicrap, Timothy},
 journal={arXiv preprint arXiv:2509.24527}, year={2025}
}

@inproceedings{oh2017vpn,
 title={Value Prediction Network}, author={Oh, Junhyuk and Singh, Satinder and Lee, Honglak},
 booktitle={Adv. Neural Inform. Process. Syst.}, volume={30}, year={2017}
}

@article{schrittwieser2020muzero,
 title={Mastering {Atari}, {Go}, Chess and Shogi by Planning with a Learned Model},
 author={Schrittwieser, Julian and Antonoglou, Ioannis and Hubert, Thomas and Simonyan, Karen and Sifre, Laurent and Schmitt, Simon and Guez, Arthur and Lockhart, Edward and Hassabis, Demis and Graepel, Thore and Lillicrap, Timothy and Silver, David},
 journal={Nature}, volume={588}, pages={604--609}, year={2020}, doi={10.1038/s41586-020-03051-4}
}

@inproceedings{parisotto2018neuralmap,
 title={{Neural Map}: Structured Memory for Deep Reinforcement Learning},
 author={Parisotto, Emilio and Salakhutdinov, Ruslan}, booktitle={Int. Conf. Learn. Represent.}, year={2018}
}

@inproceedings{chaplot2020activeslam,
 title={Learning to Explore Using {Active Neural SLAM}},
 author={Chaplot, Devendra Singh and Gandhi, Dhiraj and Gupta, Saurabh and Gupta, Abhinav and Salakhutdinov, Ruslan},
 booktitle={Int. Conf. Learn. Represent.}, year={2020}
}

@inproceedings{battaglia2016interaction,
 title={Interaction Networks for Learning about Objects, Relations and Physics},
 author={Battaglia, Peter and Pascanu, Razvan and Lai, Matthew and Rezende, Danilo Jimenez and Kavukcuoglu, Koray},
 booktitle={Adv. Neural Inform. Process. Syst.}, volume={29}, year={2016}
}

@inproceedings{byravan2017se3,
 title={{SE3-Nets}: Learning Rigid Body Motion Using Deep Neural Networks},
 author={Byravan, Arunkumar and Fox, Dieter},
 booktitle={IEEE International Conference on Robotics and Automation}, year={2017}
}

@inproceedings{kipf2020cswm,
 title={Contrastive Learning of Structured World Models},
 author={Kipf, Thomas and van der Pol, Elise and Welling, Max}, booktitle={Int. Conf. Learn. Represent.}, year={2020}
}

@article{assran2025vjepa2,
 title={{V-JEPA 2}: Self-Supervised Video Models Enable Understanding, Prediction and Planning},
 author={Assran, Mido and Bardes, Adrien and Fan, David and Garrido, Quentin and Howes, Russell and Komeili, Mojtaba and Muckley, Matthew and Rizvi, Ammar and Roberts, Claire and Sinha, Koustuv and Zholus, Artem and Arnaud, Sergio and Gejji, Abha and Martin, Ada and Hogan, Francois Robert and Dugas, Daniel and Bojanowski, Piotr and Khalidov, Vasil and Labatut, Patrick and Massa, Francisco and Szafraniec, Marc and Krishnakumar, Kapil and Li, Yong and Ma, Xiaodong and Chandar, Sarath and Meier, Franziska and LeCun, Yann and Rabbat, Michael and Ballas, Nicolas},
 journal={arXiv preprint arXiv:2506.09985}, year={2025}
}

@article{li2026dreamledger,
 title={{DreamLedger}: Where to Refuse World-Model Imagination Using Execution-Settled Credit},
 author={Li, Xianyao and Tian, Ruitong and Min, Rui and Xu, Fang and Du, Eric Jing},
 journal={arXiv preprint arXiv:2608.23863}, year={2026}
}

@inproceedings{chevalier2023miniworld,
 title={{Minigrid} \& {Miniworld}: Modular \& Customizable Reinforcement Learning Environments for Goal-Oriented Tasks},
 author={Chevalier-Boisvert, Maxime and Dai, Bolun and Towers, Mark and de Lazcano, Rodrigo and Willems, Lucas and Lahlou, Salem and Pal, Suman and Castro, Pablo Samuel and Terry, Jordan},
 booktitle={Advances in Neural Information Processing Systems, Datasets and Benchmarks Track}, volume={36}, pages={73383--73394}, year={2023}
}

@article{yuan2017gelsight,
 title={{GelSight}: High-Resolution Robot Tactile Sensors for Estimating Geometry and Force},
 author={Yuan, Wenzhen and Dong, Siyuan and Adelson, Edward H.},
 journal={Sensors}, volume={17}, number={12}, pages={2762}, year={2017}, doi={10.3390/s17122762}
}

@article{nvidia2025cosmos,
 title={Cosmos World Foundation Model Platform for Physical {AI}},
 author={Agarwal, Niket and Ali, Arslan and Bala, Maciej and Balaji, Yogesh and Barker, Erik and others},
 journal={arXiv preprint arXiv:2501.03575}, year={2025}
}

@inproceedings{lee2024dreamsmooth,
 title={{DreamSmooth}: Improving Model-based Reinforcement Learning via Reward Smoothing},
 author={Lee, Vint and Abbeel, Pieter and Lee, Youngwoon},
 booktitle={Int. Conf. Learn. Represent.}, year={2024}
}

@inproceedings{fraccaro2018gtmsm,
 title={Generative Temporal Models with Spatial Memory for Partially Observed Environments},
 author={Fraccaro, Marco and Rezende, Danilo Jimenez and Zwols, Yori and Pritzel, Alexander and Eslami, S. M. Ali and Viola, Fabio},
 booktitle={Int. Conf. Mach. Learn.}, pages={1549--1558}, year={2018}
}

@inproceedings{gupta2017cmp,
 title={Cognitive Mapping and Planning for Visual Navigation},
 author={Gupta, Saurabh and Davidson, James and Levine, Sergey and Sukthankar, Rahul and Malik, Jitendra},
 booktitle={IEEE Conf. Comput. Vis. Pattern Recog.}, pages={7272--7281}, year={2017}
}

@inproceedings{lillemark2026flowm,
 title={Flow Equivariant World Models: Structured Memory for Dynamic Environments},
 author={Lillemark, Hansen and Huang, Benhao and Zhan, Fangneng and Du, Yilun and Keller, T. Anderson},
 booktitle={Int. Conf. Mach. Learn.}, pages={73968--74002}, year={2026}
}

@inproceedings{po2025longcontext,
 title={Long-Context State-Space Video World Models},
 author={Po, Ryan and Nitzan, Yotam and Zhang, Richard and Chen, Berlin and Dao, Tri and Shechtman, Eli and Wetzstein, Gordon and Huang, Xun},
 booktitle={Int. Conf. Comput. Vis.}, pages={8733--8744}, year={2025}
}

@inproceedings{xiao2025worldmem,
 title={{WorldMem}: Long-term Consistent World Simulation with Memory},
 author={Xiao, Zeqi and Lan, Yushi and Zhou, Yifan and Ouyang, Wenqi and Yang, Shuai and Zeng, Yanhong and Pan, Xingang},
 booktitle={Adv. Neural Inform. Process. Syst.}, year={2025}
}

@inproceedings{wu2025spatialmemory,
 title={Video World Models with Long-term Spatial Memory},
 author={Wu, Tong and Yang, Shuai and Po, Ryan and Xu, Yinghao and Liu, Ziwei and Lin, Dahua and Wetzstein, Gordon},
 booktitle={Adv. Neural Inform. Process. Syst.}, year={2025}
}

@article{higuera2026vtwm,
 title={Visuo-Tactile World Models},
 author={Higuera, Carolina and Arnaud, Sergio and Boots, Byron and Mukadam, Mustafa and Hogan, Francois Robert and Meier, Franziska},
 journal={arXiv preprint arXiv:2602.06001}, year={2026}
}

@article{gao2026scandreamer,
 title={{ScanDreamer}: A Map-Grounded World Model for Quadrotor Navigation},
 author={Gao, Yunyi and Shi, Yixing and Liu, Yize and Tran, Hai-Duong and Nguyen, Nhat-Hoang P. and Wang, Shuai and Deng, Yi and Liu, Shengming and Goh, Sim Kuan and Pham, Duc-Thinh and Alam, Sameer},
 journal={arXiv preprint arXiv:2611.11605}, year={2026}
}

@article{burchi2026ttmbrl,
 title={Encoding Memory in World Models with Test-Time Model-Based Reinforcement Learning},
 author={Burchi, Maxime and Dharejo, Fayaz Ali and Timofte, Radu},
 journal={arXiv preprint arXiv:2611.09625}, year={2026}
}
\clearpage
\backrefsetup{disable}
\appendix
\makeatletter
\setlength{\@fptop}{0pt}
\setlength{\@dblfptop}{0pt}
\makeatother
\section{Implementation and memory accounting}
\label{app:method}
EPM couples explicit perceptual state to neural imagination, reward prediction and actor--critic learning. Its navigation comparisons share the perception front end across all arms. An additional policy/value-only comparison uses visual odometry without heading correction, identified separately below and in Table~\ref{tab:baselines}. This appendix specifies the state updates, learning objectives, evaluation protocols and complete results for the reported comparisons.

\begin{figure*}[t]\centering\includegraphics[width=\textwidth]{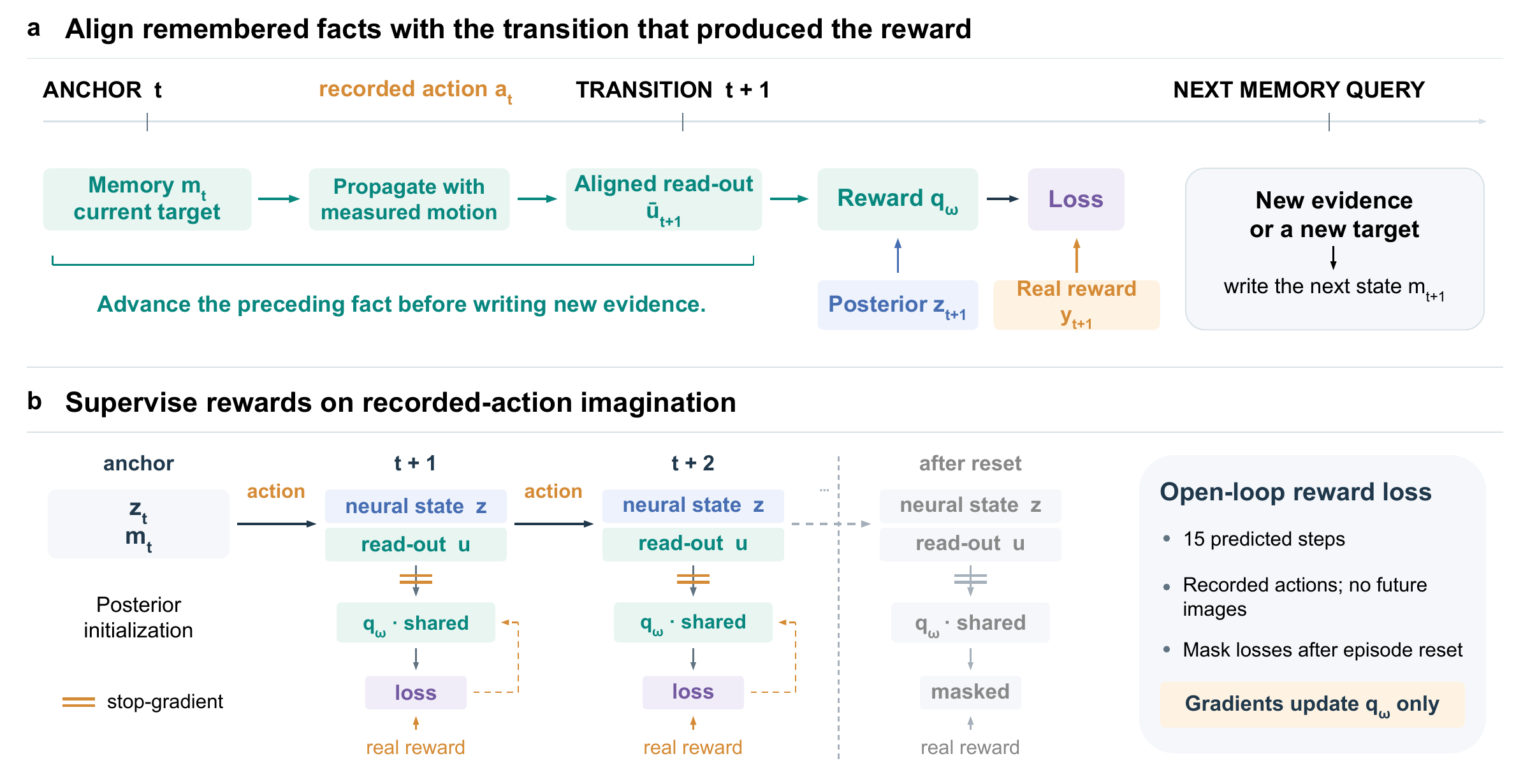}
\caption{\textbf{Training the reward interface.} (a) Posterior reward training pairs each reward with the read-out advanced from the preceding memory, not with the state stored after the next write or target switch. (b) The open-loop term rolls the prior along recorded actions, compares the shared reward head with real future rewards and masks steps after an episode reset. Double bars mark stopped gradients on features; dashed arrows show gradients into the reward head.}
\label{fig:training}\end{figure*}

\paragraph{Real-state update and local projection.}
The writer uses a calibrated detector and visual odometry to maintain seen landmarks and the estimated agent pose. A target query returns a known flag, cosine and sine of relative bearing, log-distance, and log-age. Unknown targets gate the geometric slots. The imagination initializer recovers a target vector from this query, starts the local position and yaw at zero, and carries the query age. It does not initialize an entire map rollout. For local position $p_h$, yaw $\psi_h$ and target vector $v$, a predicted translation and rotation produce
\begin{align}
 p_{h+1}&=p_h+\operatorname{Rot}(\psi_h)\widehat d_h,\\
 \psi_{h+1}&=\psi_h+\widehat\omega_h,\\
 v^{\rm rel}_{h+1}&=\operatorname{Rot}(-\psi_{h+1})(v-p_{h+1}).
\end{align}
The known flag is retained, age advances, and bearing/distance are read from $v^{\rm rel}$. The neural model continues to handle dynamics not represented by this local state. The motion head predicts displacement and yaw change from the pre-action latent feature and candidate action; its supervision is the subsequent measured motion.

\paragraph{Perception calibration.}
The detector, odometry and heading networks are fitted offline using simulator-derived labels on calibration trajectories, and frozen before policy training. At deployment they read images and actions. Memory Maze's six-color detector uses pooled $9\times9$, $11\times11$ and $13\times13$ calibration data, with a visibility half-angle of 0.75 radians; odometry is trained on pooled $9\times9$, $11\times11$ and $13\times13$ calibration walks. The heading network learns $(\cos 4\theta,\sin 4\theta)$ from a single $64\times64$ frame, exploiting axis-aligned walls. It uses 87,000 training frames and 15,000 validation frames from separate $9\times9$ calibration episodes, with 30 epochs, batch size 256 and Adam at initial learning rate $10^{-3}$ with cosine decay. After integrating odometry, the writer corrects yaw to the predicted heading on the nearest $90^\circ$ branch. No heading correction is applied inside imagination, where no images are available. MiniWorld uses a separately calibrated detector and a three-output odometry network for forward, lateral and angular displacement, trained on calibration layouts excluded from policy training and evaluation. It uses no heading correction. Each experiment shares these frozen networks across arms; the reported interaction budgets exclude this offline calibration.

\paragraph{Neural paths and parameter accounting.}
Memory Maze uses a 31-value memory observation: six target slots, eighteen detector-stream slots, five read-out slots and two motion slots. MiniWorld adds lateral motion, giving 32 values and a three-output motion head. Every arm encodes and reconstructs the complete observation; absent direct channels are zero-filled. Reward receives five additional slots, continuation retains neural input, and actor, critic, slow critic and replay value consume their assigned read-outs. The trainable runtime groups contain 12,222,441 parameters for Memory Maze and 12,223,211 for MiniWorld, identical across arms within an implementation and excluding fixed perception networks. The five reward-input slots account for 1,280 weights, present in every arm. Batches contain 16 sequences of length 64, the training ratio is 32, the imagination horizon is 15, and the return horizon is 333. Gradients from the additional open-loop reward loss update only the reward head (Fig.~\ref{fig:training}).

\paragraph{Memory storage.}
The sparse visual state's serialized payload occupies 48 bytes. The neural recurrent state separately occupies 8,224 bytes (8,192 for deterministic float32 state and 32 stochastic indices), while visual odometry retains the previous $64\times64\times3$ image, or 12,288 bytes. Network parameters and replay are additional storage components. The real-tactile memory contains 48 payload bytes and up to 9 bytes of validity and age bookkeeping.

\section{Control experiments and complete contrasts}
\label{app:formal}
\paragraph{Component design.}
Table~\ref{tab:interfaces} defines arms by their direct policy/value access, direct reward access and open-loop reward loss. Every arm receives the same memory observation; holding and evolution refer to imagination. Both reward-access arms use transition-aligned replay inputs. The access-by-evolution interaction is measured under common OL training. Reward access without OL is not a tested cell.

\paragraph{Training runs and evaluation checkpoints.}
Memory Maze $9\times9$ uses twelve initialization labels and $13\times13$ uses six, each evaluated at 250k, 500k and the 1M endpoint. MiniWorld uses twelve labels, evaluated at 1M and its 2M endpoint. Each return averages 48 fixed layout/target/spawn identities. Tables~\ref{tab:formal_full_0}--\ref{tab:formal_full_2} give all pre-specified contrasts, paired 95\% $t$ intervals, counts of positive labels and exact two-sided sign-flip sensitivity analyses; Table~\ref{tab:formal_seeds} gives label-level returns. The reward-access contrast is the designated primary analysis at $9\times9$; the remaining contrasts are pre-specified secondary analyses. Inference is separate for each experiment; shared arms make contrasts dependent. The $13\times13$ experiment trains all four non-held arms from scratch. Label-level endpoint returns are indexed within each experiment.

\begin{table*}[t]\centering\footnotesize
\caption{\textbf{Memory Maze $9\times9$: all pre-specified contrasts at every evaluation checkpoint.} $n=12$ initialization labels; $\Delta$ with paired 95\% $t$ interval, number of positive labels $n_{+}$ and exact two-sided sign-flip $p$. The endpoint is 1M.}
\label{tab:formal_full_0}
\setlength{\tabcolsep}{0pt}
\begin{tabular}{@{}l@{\hskip 7pt}r@{\hskip 3pt}l@{\hskip 4pt}c@{\hskip 4pt}r@{\hskip 8pt}r@{\hskip 3pt}l@{\hskip 4pt}c@{\hskip 4pt}r@{\hskip 8pt}r@{\hskip 3pt}l@{\hskip 4pt}c@{\hskip 4pt}r@{}}\toprule
 & \multicolumn{4}{c}{0.25M} & \multicolumn{4}{c}{0.5M} & \multicolumn{4}{c}{1M} \\
Contrast & \multicolumn{1}{c}{$\Delta$} & \multicolumn{1}{c}{95\% CI} & $n_{+}$ & \multicolumn{1}{c}{$p$} & \multicolumn{1}{c}{$\Delta$} & \multicolumn{1}{c}{95\% CI} & $n_{+}$ & \multicolumn{1}{c}{$p$} & \multicolumn{1}{c}{$\Delta$} & \multicolumn{1}{c}{95\% CI} & $n_{+}$ & \multicolumn{1}{c}{$p$} \\
\midrule
EPM $-$ Evolved+OL & $+5.85$ & [$4.77$, $6.94$] & 12/12 & 0.0005 & $+5.90$ & [$3.94$, $7.86$] & 11/12 & 0.0010 & $+5.51$ & [$4.46$, $6.57$] & 12/12 & 0.0005 \\
EPM $-$ Obs+OL & $+8.81$ & [$7.92$, $9.69$] & 12/12 & 0.0005 & $+12.63$ & [$10.64$, $14.62$] & 12/12 & 0.0005 & $+7.89$ & [$6.17$, $9.60$] & 12/12 & 0.0005 \\
EPM $-$ Held+R+OL & $+8.20$ & [$7.08$, $9.33$] & 12/12 & 0.0005 & $+12.07$ & [$10.24$, $13.89$] & 12/12 & 0.0005 & $+13.12$ & [$12.01$, $14.23$] & 12/12 & 0.0005 \\
Evolved+OL $-$ Held+OL & $+1.99$ & [$0.91$, $3.07$] & 11/12 & 0.0044 & $+4.61$ & [$3.24$, $5.98$] & 12/12 & 0.0005 & $+2.51$ & [$1.36$, $3.66$] & 11/12 & 0.0015 \\
Held+R+OL $-$ Held+OL & $-0.36$ & [$-0.95$, $0.24$] & 4/12 & 0.2153 & $-1.55$ & [$-3.19$, $0.08$] & 2/12 & 0.0635 & $-5.10$ & [$-6.34$, $-3.86$] & 0/12 & 0.0005 \\
Evolved+OL $-$ Obs+OL & $+2.95$ & [$2.20$, $3.70$] & 12/12 & 0.0005 & $+6.73$ & [$4.86$, $8.59$] & 12/12 & 0.0005 & $+2.37$ & [$0.21$, $4.54$] & 7/12 & 0.0366 \\
Held+OL $-$ Obs+OL & $+0.96$ & [$0.21$, $1.71$] & 10/12 & 0.0200 & $+2.12$ & [$0.53$, $3.70$] & 10/12 & 0.0176 & $-0.14$ & [$-2.02$, $1.74$] & 7/12 & 0.8721 \\
Held+R+OL $-$ Obs+OL & $+0.60$ & [$0.08$, $1.12$] & 9/12 & 0.0244 & $+0.56$ & [$-1.25$, $2.38$] & 6/12 & 0.5005 & $-5.24$ & [$-6.68$, $-3.79$] & 0/12 & 0.0005 \\
EPM $-$ Evolved & $+3.52$ & [$1.50$, $5.53$] & 11/12 & 0.0024 & $+4.32$ & [$2.46$, $6.18$] & 11/12 & 0.0015 & $+6.08$ & [$4.46$, $7.71$] & 12/12 & 0.0005 \\
Evolved+OL $-$ Evolved & $-2.34$ & [$-4.10$, $-0.57$] & 2/12 & 0.0098 & $-1.58$ & [$-3.22$, $0.07$] & 3/12 & 0.0591 & $+0.57$ & [$-1.55$, $2.70$] & 7/12 & 0.5659 \\
Evolved $-$ Obs+OL & $+5.29$ & [$3.40$, $7.18$] & 12/12 & 0.0005 & $+8.31$ & [$6.70$, $9.91$] & 12/12 & 0.0005 & $+1.80$ & [$-0.38$, $3.98$] & 11/12 & 0.0967 \\
Reward access $\times$ evolution & $+6.21$ & [$5.22$, $7.20$] & 12/12 & 0.0005 & $+7.45$ & [$5.19$, $9.72$] & 12/12 & 0.0005 & $+10.61$ & [$9.35$, $11.86$] & 12/12 & 0.0005 \\
\bottomrule\end{tabular}\end{table*}
\begin{table*}[t]\centering\footnotesize
\caption{\textbf{Memory Maze $13\times13$: all pre-specified contrasts at every evaluation checkpoint.} $n=6$ initialization labels; $\Delta$ with paired 95\% $t$ interval, number of positive labels $n_{+}$ and exact two-sided sign-flip $p$. The endpoint is 1M.}
\label{tab:formal_full_1}
\setlength{\tabcolsep}{0pt}
\begin{tabular}{@{}l@{\hskip 7pt}r@{\hskip 3pt}l@{\hskip 4pt}c@{\hskip 4pt}r@{\hskip 8pt}r@{\hskip 3pt}l@{\hskip 4pt}c@{\hskip 4pt}r@{\hskip 8pt}r@{\hskip 3pt}l@{\hskip 4pt}c@{\hskip 4pt}r@{}}\toprule
 & \multicolumn{4}{c}{0.25M} & \multicolumn{4}{c}{0.5M} & \multicolumn{4}{c}{1M} \\
Contrast & \multicolumn{1}{c}{$\Delta$} & \multicolumn{1}{c}{95\% CI} & $n_{+}$ & \multicolumn{1}{c}{$p$} & \multicolumn{1}{c}{$\Delta$} & \multicolumn{1}{c}{95\% CI} & $n_{+}$ & \multicolumn{1}{c}{$p$} & \multicolumn{1}{c}{$\Delta$} & \multicolumn{1}{c}{95\% CI} & $n_{+}$ & \multicolumn{1}{c}{$p$} \\
\midrule
EPM $-$ Evolved+OL & $+1.76$ & [$-1.52$, $5.05$] & 4/6 & 0.3125 & $+9.13$ & [$4.56$, $13.70$] & 6/6 & 0.0312 & $+11.39$ & [$7.89$, $14.89$] & 6/6 & 0.0312 \\
EPM $-$ Evolved & $+1.37$ & [$-0.96$, $3.70$] & 5/6 & 0.2188 & $+0.83$ & [$-4.93$, $6.58$] & 3/6 & 0.7500 & $+7.00$ & [$3.80$, $10.21$] & 6/6 & 0.0312 \\
EPM $-$ Obs+OL & $+3.71$ & [$1.67$, $5.75$] & 6/6 & 0.0312 & $+14.92$ & [$12.00$, $17.85$] & 6/6 & 0.0312 & $+16.19$ & [$12.67$, $19.72$] & 6/6 & 0.0312 \\
Evolved+OL $-$ Evolved & $-0.39$ & [$-2.81$, $2.03$] & 3/6 & 0.8125 & $-8.30$ & [$-15.85$, $-0.76$] & 1/6 & 0.0625 & $-4.39$ & [$-8.16$, $-0.61$] & 0/6 & 0.0312 \\
Evolved+OL $-$ Obs+OL & $+1.95$ & [$-0.36$, $4.26$] & 4/6 & 0.1250 & $+5.80$ & [$0.34$, $11.25$] & 6/6 & 0.0312 & $+4.81$ & [$0.72$, $8.89$] & 6/6 & 0.0312 \\
Evolved $-$ Obs+OL & $+2.34$ & [$0.56$, $4.11$] & 6/6 & 0.0312 & $+14.10$ & [$7.98$, $20.22$] & 6/6 & 0.0312 & $+9.19$ & [$6.69$, $11.69$] & 6/6 & 0.0312 \\
\bottomrule\end{tabular}\end{table*}
\begin{table*}[t]\centering\footnotesize
\caption{\textbf{MiniWorld $9\times9$: all pre-specified contrasts at every evaluation checkpoint.} $n=12$ initialization labels; $\Delta$ with paired 95\% $t$ interval, number of positive labels $n_{+}$ and exact two-sided sign-flip $p$. The endpoint is 2M.}
\label{tab:formal_full_2}
\setlength{\tabcolsep}{0pt}
\begin{tabular}{@{}l@{\hskip 16pt}r@{\hskip 5pt}l@{\hskip 8pt}c@{\hskip 8pt}r@{\hskip 22pt}r@{\hskip 5pt}l@{\hskip 8pt}c@{\hskip 8pt}r@{}}\toprule
 & \multicolumn{4}{c}{1M} & \multicolumn{4}{c}{2M} \\
Contrast & \multicolumn{1}{c}{$\Delta$} & \multicolumn{1}{c}{95\% CI} & $n_{+}$ & \multicolumn{1}{c}{$p$} & \multicolumn{1}{c}{$\Delta$} & \multicolumn{1}{c}{95\% CI} & $n_{+}$ & \multicolumn{1}{c}{$p$} \\
\midrule
EPM $-$ Evolved+OL & $+2.18$ & [$0.68$, $3.68$] & 10/12 & 0.0098 & $+1.00$ & [$-0.04$, $2.03$] & 10/12 & 0.0620 \\
EPM $-$ Obs+OL & $+4.30$ & [$2.93$, $5.66$] & 12/12 & 0.0005 & $+1.09$ & [$0.02$, $2.17$] & 9/12 & 0.0493 \\
EPM $-$ Held+R+OL & $+16.89$ & [$15.43$, $18.36$] & 12/12 & 0.0005 & $+18.46$ & [$16.04$, $20.89$] & 12/12 & 0.0005 \\
Evolved+OL $-$ Held+OL & $+3.90$ & [$3.01$, $4.78$] & 12/12 & 0.0005 & $+2.31$ & [$1.10$, $3.52$] & 11/12 & 0.0034 \\
Held+R+OL $-$ Held+OL & $-10.82$ & [$-12.52$, $-9.11$] & 0/12 & 0.0005 & $-15.15$ & [$-17.95$, $-12.36$] & 0/12 & 0.0005 \\
Evolved+OL $-$ Obs+OL & $+2.12$ & [$0.93$, $3.30$] & 11/12 & 0.0039 & $+0.09$ & [$-0.84$, $1.03$] & 5/12 & 0.8281 \\
Held+OL $-$ Obs+OL & $-1.78$ & [$-2.90$, $-0.66$] & 1/12 & 0.0044 & $-2.22$ & [$-3.63$, $-0.81$] & 1/12 & 0.0039 \\
Held+R+OL $-$ Obs+OL & $-12.60$ & [$-14.37$, $-10.82$] & 0/12 & 0.0005 & $-17.37$ & [$-19.58$, $-15.16$] & 0/12 & 0.0005 \\
EPM $-$ Evolved & $+7.68$ & [$6.33$, $9.03$] & 12/12 & 0.0005 & $+3.74$ & [$2.51$, $4.97$] & 12/12 & 0.0005 \\
Evolved+OL $-$ Evolved & $+5.50$ & [$3.93$, $7.07$] & 12/12 & 0.0005 & $+2.74$ & [$1.16$, $4.33$] & 11/12 & 0.0044 \\
Evolved $-$ Obs+OL & $-3.38$ & [$-5.04$, $-1.73$] & 1/12 & 0.0010 & $-2.65$ & [$-4.26$, $-1.04$] & 3/12 & 0.0059 \\
Reward access $\times$ evolution & $+13.00$ & [$11.23$, $14.76$] & 12/12 & 0.0005 & $+16.15$ & [$13.05$, $19.25$] & 12/12 & 0.0005 \\
\bottomrule\end{tabular}\end{table*}

\begin{table*}[t]\centering\footnotesize
\caption{\textbf{Label-level mean returns at the endpoint} (48 evaluation identities per label). Labels are indexed within each experiment.}
\label{tab:formal_seeds}
\setlength{\tabcolsep}{3.2pt}
\begin{tabular}[t]{@{}lrrrrrr@{}}\toprule
\multicolumn{7}{@{}l}{Memory Maze $9\times9$, 1M}\\
\midrule
Label & Obs+OL & Held+OL & Held+R+OL & Evolved & Evolved+OL & EPM \\
\midrule
1 & 17.50 & 16.17 & 11.54 & 20.35 & 16.31 & 24.88 \\
2 & 16.56 & 17.38 & 9.75 & 18.56 & 21.56 & 25.73 \\
3 & 20.81 & 17.54 & 14.23 & 12.77 & 19.60 & 25.25 \\
4 & 19.67 & 16.23 & 11.25 & 21.17 & 20.75 & 24.06 \\
5 & 15.12 & 19.29 & 11.38 & 18.96 & 21.33 & 25.38 \\
6 & 18.77 & 13.79 & 10.04 & 21.23 & 16.60 & 24.56 \\
7 & 19.83 & 20.15 & 13.65 & 20.21 & 19.48 & 25.40 \\
8 & 13.35 & 17.94 & 10.65 & 18.83 & 20.25 & 25.58 \\
9 & 18.29 & 16.54 & 14.62 & 19.88 & 17.75 & 24.19 \\
10 & 15.15 & 17.25 & 13.73 & 19.96 & 20.08 & 26.56 \\
11 & 15.88 & 16.02 & 12.35 & 17.35 & 21.85 & 25.50 \\
12 & 16.90 & 17.85 & 11.81 & 20.19 & 20.73 & 25.38 \\
Mean & 17.32 & 17.18 & 12.08 & 19.12 & 19.69 & 25.20 \\
\bottomrule\end{tabular}\hfill
\begin{tabular}[t]{@{}lrrrrrr@{}}\toprule
\multicolumn{7}{@{}l}{MiniWorld $9\times9$, 2M}\\
\midrule
Label & Obs+OL & Held+OL & Held+R+OL & Evolved & Evolved+OL & EPM \\
\midrule
1 & 19.29 & 18.79 & 4.38 & 15.52 & 21.46 & 19.52 \\
2 & 21.48 & 17.90 & 1.73 & 15.83 & 20.12 & 19.65 \\
3 & 21.31 & 18.15 & 2.12 & 21.71 & 22.02 & 23.04 \\
4 & 20.56 & 19.77 & 1.83 & 17.12 & 22.60 & 22.71 \\
5 & 21.27 & 16.29 & 1.65 & 18.56 & 20.77 & 22.08 \\
6 & 18.54 & 16.98 & 2.85 & 17.54 & 19.23 & 22.85 \\
7 & 21.23 & 19.94 & 1.65 & 17.12 & 20.90 & 23.60 \\
8 & 21.15 & 15.04 & 4.25 & 17.23 & 19.71 & 20.90 \\
9 & 21.40 & 17.00 & 13.48 & 19.92 & 21.08 & 21.52 \\
10 & 20.48 & 20.48 & 1.83 & 20.81 & 18.69 & 22.17 \\
11 & 22.77 & 21.25 & 1.94 & 15.48 & 21.71 & 22.00 \\
12 & 19.40 & 20.65 & 2.71 & 20.25 & 21.71 & 21.94 \\
Mean & 20.74 & 18.52 & 3.37 & 18.09 & 20.83 & 21.83 \\
\bottomrule\end{tabular}

\vspace{8pt}
\begin{tabular}[t]{@{}lrrrr@{}}\toprule
\multicolumn{5}{@{}l}{Memory Maze $13\times13$, 1M}\\
\midrule
Label & Obs+OL & Evolved & Evolved+OL & EPM \\
\midrule
1 & 11.04 & 23.98 & 22.62 & 32.29 \\
2 & 14.54 & 20.98 & 18.12 & 30.73 \\
3 & 16.69 & 26.56 & 17.35 & 31.67 \\
4 & 15.98 & 25.12 & 20.38 & 26.92 \\
5 & 14.19 & 24.12 & 16.21 & 31.67 \\
6 & 12.62 & 19.44 & 19.21 & 28.96 \\
Mean & 14.18 & 23.37 & 18.98 & 30.37 \\
\bottomrule\end{tabular}
\end{table*}

\paragraph{MiniWorld endpoint.}
MiniWorld's reward-access difference is $2.18$ [$0.68$, $3.68$] at 1M and $1.00$ [$-0.04$, $2.03$] at the pre-specified 2M endpoint. At 2M, Evolved+OL minus Observation+OL is $0.09$ [$-0.84$, $1.03$], while Evolved+OL minus Held+OL is $2.31$ [$1.10$, $3.52$]. The pre-specified joint criterion, which requires EPM to exceed both Evolved+OL and Observation+OL with resolved intervals, is not met in MiniWorld because the first difference is unresolved; the difference from Observation+OL is $1.09$ [$0.02$, $2.17$]. The complete-method contrast against Evolved is a separate pre-specified comparison.

\section{Reward-anticipation evaluation}
\label{app:mechanism}
The evaluation filters each recorded episode with real observations and actions, then rolls the prior along recorded continuations for 15 steps. Predictions average four stochastic rollouts. Every tested arm reads common histories from the EPM and Evolved+OL source policies at the endpoint; estimates aggregate windows within episodes and give the two sources equal weight. VIS denotes a truly visible target at the anchor. MEM denotes a non-visible target that is known in memory, with bearing error at most 0.5 radians. A MEM event satisfies RA when the stepped read-out anticipates arrival: at the step before the first reward, its predicted distance is at most the task's reach distance plus one forward step (0.2 cells). RA uses each tested model's predicted motion, so MEM-and-RA subsets can differ across arms even on the same histories. The paired contrasts compare conditional anticipation scores, not pointwise errors on identical events. Simulator truth defines diagnostic strata and is not policy input. Later rewards after a target switch are excluded from the first-arrival contrast.

For a first reward at horizon index $k$, event mass sums predicted reward at $k-1$ through $k+1$. Non-arrival mass is the predicted horizon total scaled to three steps. Their difference is net anticipation $A$. Eligible horizon bins (1--5, 6--10, 11--15) have at least five distinct events in VIS and the compared memory set and receive equal weight. The assessability rule requires twenty distinct VIS and twenty MEM-and-RA events per cell; all twelve labels are assessable in each experiment, with no exclusions. $A$ can exceed one and is neither a probability nor an accuracy. Table~\ref{tab:mechanism_full} gives all-arm levels; the pre-specified paired contrast is EPM minus Evolved+OL for $A_{\rm MEM\cap RA}$. Secondary differences in $A_{\rm VIS}$ are $0.196$ [$0.148$, $0.244$] in Memory Maze and $0.278$ [$0.215$, $0.340$] in MiniWorld. The corresponding increases in the per-label ratio $A_{\rm MEM\cap RA}/A_{\rm VIS}$ are $0.547$ [$0.495$, $0.599$] and $0.146$ [$0.097$, $0.195$]. Ratios accompany absolute levels rather than replacing them.
\begin{table*}[t]\centering\small
\caption{\textbf{Reward-anticipation levels.} Means with 95\% $t$ intervals over twelve labels. $D=A_{\rm VIS}-A_{\rm MEM\cap RA}$. Paired contrasts between EPM and Evolved+OL are given in the text.}
\label{tab:mechanism_full}
\setlength{\tabcolsep}{8pt}
\begin{tabular}{@{}lccc@{}}\toprule
Arm & $A_{\rm VIS}$ & $A_{\rm MEM\cap RA}$ & $D$ \\
\midrule
\multicolumn{4}{@{}l}{\textit{Memory Maze $9\times9$, 1M}} \\
Observation+OL & $0.351$ [$0.307$, $0.394$] & $0.055$ [$0.035$, $0.075$] & $0.295$ [$0.261$, $0.329$] \\
Held+OL & $0.390$ [$0.348$, $0.432$] & $0.063$ [$0.048$, $0.077$] & $0.328$ [$0.289$, $0.366$] \\
Held+R+OL & $0.439$ [$0.404$, $0.474$] & $0.146$ [$0.120$, $0.172$] & $0.293$ [$0.262$, $0.323$] \\
Evolved & $0.454$ [$0.405$, $0.503$] & $0.053$ [$0.041$, $0.065$] & $0.401$ [$0.353$, $0.449$] \\
Evolved+OL & $0.482$ [$0.458$, $0.507$] & $0.154$ [$0.127$, $0.180$] & $0.329$ [$0.298$, $0.360$] \\
\textbf{EPM} & $0.679$ [$0.631$, $0.727$] & $0.587$ [$0.539$, $0.636$] & $0.092$ [$0.064$, $0.119$] \\
\midrule
\multicolumn{4}{@{}l}{\textit{MiniWorld $9\times9$, 2M}} \\
Observation+OL & $0.461$ [$0.420$, $0.502$] & $0.265$ [$0.232$, $0.298$] & $0.196$ [$0.167$, $0.225$] \\
Held+OL & $0.423$ [$0.377$, $0.468$] & $0.222$ [$0.180$, $0.264$] & $0.201$ [$0.175$, $0.227$] \\
Held+R+OL & $0.454$ [$0.378$, $0.531$] & $0.290$ [$0.232$, $0.348$] & $0.164$ [$0.133$, $0.195$] \\
Evolved & $0.556$ [$0.509$, $0.602$] & $0.277$ [$0.250$, $0.303$] & $0.279$ [$0.245$, $0.313$] \\
Evolved+OL & $0.582$ [$0.527$, $0.638$] & $0.447$ [$0.403$, $0.490$] & $0.136$ [$0.106$, $0.165$] \\
\textbf{EPM} & $0.860$ [$0.794$, $0.926$] & $0.787$ [$0.716$, $0.859$] & $0.073$ [$0.041$, $0.105$] \\
\bottomrule\end{tabular}
\end{table*}

\section{MiniWorld implementation}
\label{app:versions}
The MiniWorld task uses the maze layout generator, target identities and action budget of the navigation protocol, with its own renderer and wall-sliding kinematics. Each environment maintains an independent scene display list. Planar odometry includes forward, lateral and angular displacement. All arms use independent training runs from scratch, matched initialization within each label, and the same 1,000-action episode budget.

\section{Policy/value-only memory interface}
\label{app:fresh}
This comparison uses six initialization labels, common initial weights and identical parameter counts across the evolved, held and observation-only arms. Visual odometry without heading correction supplies the same memory observations to all arms; direct read-outs enter policy and value heads, and reward training uses posterior neural features. Evaluation pairs the same 48 maze identities across arms and masking modes. Tables~\ref{tab:new_three_arm}--\ref{tab:new_three_arm_seeds} give all checkpoints and per-label outcomes. Paired $t$ intervals and sign-flip sensitivity analyses use training labels as the unit. The primary sequence tests evolved against observation-only at 500k, then 1M; other contrasts are secondary and unadjusted.

\begin{table*}[t]
\centering\small
\caption{\textbf{Policy/value-only comparison on Memory Maze $13\times13$.} Six initialization labels, 48 identity-paired evaluation mazes per label. Return differences and 95\% paired $t$ intervals use labels as the unit; $p_{\mathrm{flip}}$ is the exact two-sided sign-flip sensitivity analysis. The primary sequence tests Evolved versus observation-only at 500k, then 1M. All other rows are secondary. $^{\dagger}$The $t$ interval and sign-flip test disagree at the 0.05 level. Shared arms make the contrasts correlated.}
\label{tab:new_three_arm}
\begin{tabular}{llrccc}\toprule
Comparison & Steps & Return difference [95\% CI] & Positive & $p_{\mathrm{flip}}$ & Role \\
\midrule
Evolved $-$ observation-only & 250k & +0.60 [$-$0.74, +1.94] & 5/6 & 0.34375 & secondary \\
Evolved $-$ observation-only & 500k & +2.72 [+0.61, +4.83] & 6/6 & 0.03125 & primary \\
Evolved $-$ observation-only & 1M & +5.24 [+1.05, +9.43] & 6/6 & 0.03125 & primary \\
\midrule
Evolved $-$ held & 250k & +0.83 [$-$0.69, +2.36] & 4/6 & 0.1875 & secondary \\
Evolved $-$ held & 500k & +2.17 [$-$0.44, +4.77] & 6/6 & 0.03125$^{\dagger}$ & secondary \\
Evolved $-$ held & 1M & +5.35 [+0.31, +10.40] & 5/6 & 0.0625$^{\dagger}$ & secondary \\
\midrule
Held $-$ observation-only & 250k & $-$0.23 [$-$0.60, +0.13] & 1/6 & 0.15625 & secondary \\
Held $-$ observation-only & 500k & +0.56 [$-$0.41, +1.52] & 4/6 & 0.1875 & secondary \\
Held $-$ observation-only & 1M & $-$0.12 [$-$3.00, +2.77] & 3/6 & 0.84375 & secondary \\
\bottomrule\end{tabular}\end{table*}

\begin{table*}[t]\centering\small
\caption{Read-out dependence differences in the policy/value-only comparison (descriptive). Dependence is the intact-minus-masked return within an arm. Intervals are paired $t$ intervals over six labels; sign-flip $p$ values are sensitivity analyses. Masking affects the observation read-out and, where present, its direct actor/critic path. These differences share intact returns with Table~\ref{tab:new_three_arm} and are not independent evidence.}
\label{tab:new_dependence}
\begin{tabular}{llrcc}\toprule
Comparison & Steps & Dependence difference [95\% CI] & Positive & $p_{\mathrm{flip}}$ \\
\midrule
Evolved $-$ observation-only & 250k & +0.76 [$-$0.05, +1.56] & 6/6 & 0.03125 \\
Evolved $-$ observation-only & 500k & +1.57 [$-$0.24, +3.38] & 5/6 & 0.0625 \\
Evolved $-$ observation-only & 1M & +2.59 [+0.52, +4.67] & 6/6 & 0.03125 \\
\midrule
Evolved $-$ held & 250k & +0.82 [$-$0.05, +1.69] & 6/6 & 0.03125 \\
Evolved $-$ held & 500k & +1.46 [$-$0.63, +3.54] & 5/6 & 0.09375 \\
Evolved $-$ held & 1M & +1.82 [$-$0.65, +4.29] & 4/6 & 0.125 \\
\midrule
Held $-$ observation-only & 250k & $-$0.06 [$-$0.18, +0.06] & 2/6 & 0.28125 \\
Held $-$ observation-only & 500k & +0.12 [$-$0.31, +0.54] & 3/6 & 0.5 \\
Held $-$ observation-only & 1M & +0.77 [$-$0.32, +1.87] & 5/6 & 0.125 \\
\bottomrule\end{tabular}\end{table*}

\begin{table*}[t]\centering\scriptsize
\caption{Policy/value-only comparison, all initialization labels and budgets. Returns are means over 48 mazes. The three contrasts compare intact returns; dependence is intact minus masked return. O denotes observation-only and F denotes held.}
\label{tab:new_three_arm_seeds}
\begin{tabular}{lrrrrrrrrr}\toprule
Label & Evolved & F & O & Evolved $-$ O & Evolved $-$ F & F $-$ O & Dep. Evolved & Dep. F & Dep. O \\
\midrule\multicolumn{10}{l}{\textit{250k}} \\
1 & 1.229 & 0.250 & 0.625 & +0.604 & +0.979 & $-$0.375 & 0.708 & 0.000 & $-$0.083 \\
2 & 0.729 & 0.458 & 0.542 & +0.188 & +0.271 & $-$0.083 & 0.229 & $-$0.021 & 0.021 \\
3 & 3.875 & 0.250 & 0.854 & +3.021 & +3.625 & $-$0.604 & 2.417 & $-$0.021 & 0.167 \\
4 & 0.583 & 0.896 & 0.562 & +0.021 & $-$0.312 & +0.333 & 0.354 & 0.083 & 0.021 \\
5 & 0.562 & 0.750 & 1.292 & $-$0.729 & $-$0.188 & $-$0.542 & 0.312 & $-$0.083 & 0.042 \\
6 & 1.188 & 0.562 & 0.688 & +0.500 & +0.625 & $-$0.125 & 0.729 & $-$0.104 & 0.042 \\
\midrule\multicolumn{10}{l}{\textit{500k}} \\
1 & 8.375 & 1.438 & 1.979 & +6.396 & +6.938 & $-$0.542 & 5.062 & 0.042 & 0.167 \\
2 & 2.792 & 2.500 & 2.250 & +0.542 & +0.292 & +0.250 & 0.604 & 0.375 & $-$0.125 \\
3 & 4.688 & 1.896 & 1.958 & +2.729 & +2.792 & $-$0.062 & 1.375 & 0.208 & 0.125 \\
4 & 3.250 & 2.375 & 1.688 & +1.562 & +0.875 & +0.688 & 1.458 & 0.292 & 0.292 \\
5 & 4.250 & 3.292 & 1.167 & +3.083 & +0.958 & +2.125 & $-$0.104 & 0.688 & 0.021 \\
6 & 3.979 & 2.833 & 1.958 & +2.021 & +1.146 & +0.875 & 1.625 & $-$0.312 & 0.125 \\
\midrule\multicolumn{10}{l}{\textit{1M}} \\
1 & 14.417 & 8.021 & 6.917 & +7.500 & +6.396 & +1.104 & 4.479 & 2.104 & $-$0.104 \\
2 & 8.146 & 3.208 & 7.583 & +0.562 & +4.938 & $-$4.375 & 0.188 & 1.125 & $-$0.042 \\
3 & 12.250 & 2.750 & 2.458 & +9.792 & +9.500 & +0.292 & 4.542 & $-$0.271 & 0.688 \\
4 & 9.292 & 3.125 & 3.521 & +5.771 & +6.167 & $-$0.396 & 4.500 & 1.312 & 0.646 \\
5 & 9.500 & 13.292 & 9.333 & +0.167 & $-$3.792 & +3.958 & 3.083 & 4.167 & 3.062 \\
6 & 13.417 & 4.500 & 5.771 & +7.646 & +8.917 & $-$1.271 & 4.438 & 1.854 & 1.417 \\
\bottomrule\end{tabular}\end{table*}

Masking removes the target read-out from the observation vector and its direct path while retaining the remaining observation history and neural state. At 1M, the evolved arm's additional masking loss over observation-only is positive under both the paired $t$ interval and sign-flip analysis. Its dependence contrast against held remains unresolved by the $t$ interval (Table~\ref{tab:new_dependence}).

\section{Alternative state updates}
\label{app:historical}
\paragraph{Advancing age versus geometry.}
A clock-only comparison, on the front end without heading correction, increments memory age during imagination while holding bearing and distance at the anchor. Table~\ref{tab:clock_seeds} reports evolved, held and clock-only arms at all three budgets on six initialization labels and 48 common evaluation mazes. At 1M, evolved geometry exceeds clock-only on all six labels. The evolved--held difference is positive on five labels, with a seed-level bootstrap interval crossing zero. Thus this comparison distinguishes spatial updates from merely advancing the clock, while the main six-arm experiments evaluate evolution under the reward-learning design.

\begin{table*}[t]\centering\footnotesize
\caption{Clock-only control on $13\times13$ without heading correction: six initialization labels and 48 common evaluation mazes. E = evolved geometry; H = held; C = clock-only (age advances, geometry is held). Returns and read-out dependence are listed as E / H / C. Within-label differences use paired-maze bootstrap 95\% intervals; dependence is intact minus masked return.}
\label{tab:clock_seeds}
\setlength{\tabcolsep}{7pt}
\begin{tabular}{@{}cllll@{}}\toprule
Label & Return (E / H / C) & E $-$ C [95\% CI] & C $-$ H [95\% CI] & Dependence (E / H / C) \\
\midrule\multicolumn{5}{l}{\textit{250k}} \\
1 & 0.8 / 0.8 / 0.9 & $-$0.08 [$-$0.58, +0.48] & +0.12 [$-$0.21, +0.46] & +0.15 / +0.10 / $-$0.04 \\
2 & 2.1 / 1.5 / 0.6 & +1.52 [+0.81, +2.35] & $-$0.90 [$-$1.35, $-$0.46] & +0.90 / $-$0.10 / 0.00 \\
3 & 2.1 / 0.8 / 0.4 & +1.73 [+1.19, +2.33] & $-$0.48 [$-$0.71, $-$0.25] & +0.83 / 0.00 / +0.06 \\
4 & 1.1 / 1.1 / 0.5 & +0.65 [+0.33, +0.96] & $-$0.60 [$-$0.94, $-$0.27] & +0.42 / +0.21 / +0.10 \\
5 & 1.7 / 1.0 / 1.6 & +0.12 [$-$0.46, +0.71] & +0.54 [+0.02, +1.02] & +1.25 / +0.04 / +0.27 \\
6 & 3.3 / 1.0 / 0.8 & +2.48 [+1.65, +3.46] & $-$0.21 [$-$0.58, +0.12] & +1.60 / $-$0.06 / +0.19 \\
\midrule\multicolumn{5}{l}{\textit{500k}} \\
1 & 5.9 / 4.6 / 2.2 & +3.65 [+2.10, +5.29] & $-$2.38 [$-$3.21, $-$1.54] & +2.25 / $-$0.21 / +0.52 \\
2 & 6.2 / 1.5 / 1.6 & +4.52 [+3.08, +6.04] & +0.17 [$-$0.46, +0.77] & +2.40 / +0.38 / +0.12 \\
3 & 9.0 / 1.6 / 3.1 & +5.88 [+4.19, +7.67] & +1.56 [+0.94, +2.17] & +5.00 / +0.19 / 0.00 \\
4 & 4.8 / 3.6 / 2.9 & +1.90 [+0.54, +3.42] & $-$0.69 [$-$1.56, +0.21] & +2.06 / $-$0.29 / +0.73 \\
5 & 8.7 / 5.1 / 1.6 & +7.12 [+5.25, +9.23] & $-$3.56 [$-$4.44, $-$2.75] & +3.10 / +1.29 / $-$0.08 \\
6 & 8.9 / 2.7 / 1.4 & +7.48 [+5.62, +9.42] & $-$1.33 [$-$1.96, $-$0.71] & +4.50 / +0.15 / $-$0.04 \\
\midrule\multicolumn{5}{l}{\textit{1M}} \\
1 & 8.2 / 13.8 / 2.9 & +5.27 [+3.00, +7.77] & $-$10.88 [$-$13.19, $-$8.60] & +2.65 / +5.40 / +0.77 \\
2 & 10.0 / 5.2 / 7.2 & +2.81 [+0.54, +5.08] & +2.02 [+0.48, +3.69] & +4.42 / +1.40 / +1.67 \\
3 & 18.5 / 5.6 / 6.7 & +11.81 [+8.10, +15.46] & +1.15 [$-$0.40, +2.83] & +8.79 / +1.69 / +1.33 \\
4 & 18.9 / 13.4 / 6.6 & +12.33 [+8.79, +15.96] & $-$6.81 [$-$8.35, $-$5.27] & +9.00 / +4.40 / +2.81 \\
5 & 13.6 / 13.0 / 4.7 & +8.83 [+6.42, +11.42] & $-$8.23 [$-$10.33, $-$6.08] & +4.77 / +5.10 / +1.65 \\
6 & 16.1 / 13.8 / 5.7 & +10.33 [+7.33, +13.42] & $-$8.08 [$-$10.62, $-$5.67] & +7.67 / +5.44 / +2.17 \\
\bottomrule\end{tabular}
\end{table*}

\section{Shared-history branch diagnostics}
\label{app:branches}
Using common histories from the policy/value-only comparison, we measure posterior read-out reconstruction and the ranking of imagined branch consequences. At 1M, median current-bearing errors are 0.325, 0.339 and 0.309 radians (evolved, held, observation-only). Explicit branch ranking agreement is 0.534, 0.521 and 0.529, compared with 0.221, 0.174 and 0.251 for decoded read-outs; the tie reference is 0.170. The explicit state thus ranks branch consequences well above the decoded read-outs.

\section{External baselines and budgets}
\label{app:baselines}
The DreamerV3 and R2I baselines use their official implementations with 16 environments and a 2M-transition replay capacity. The R2I perception variant receives the same observation stream as the policy/value-only comparison; its state-equipped variant additionally receives maintained memory through observation encoding. External-method results use three training seeds. Table~\ref{tab:baselines} separates this front end from the heading-corrected one used in the main component comparisons.

R2I returns at 1M/2M/3M are 0.778/3.326/6.743 without the added perception stream, 0.951/4.312/6.076 with it, and 0.840/3.472/9.361 with perception and maintained state. The six-label policy/value-only agent at 1M exceeds R2I with the front end at 3M by 5.09 [0.92, 9.27] under a training-seed Welch interval. Against the stronger state-equipped variant, the difference is 1.81 [$-6.50$, 10.12]. These cross-method comparisons are descriptive; the EPM component effects use the matched controls in Table~\ref{tab:main_contrasts}.

\section{Symbolic and tactile experiments}
\label{app:tactile}
\paragraph{Concentration.}
The rule simulator advances card observations, phase, reward and termination according to the public rules. Known identities come from public history; unknown identities are sampled. The remembered-fact and resampled-fact learners and the real-training reference share policy architecture, learner and iteration count. Each deployed policy is evaluated for 256 episodes.

\paragraph{Tactile branch evaluation.}
At lost-contact anchors, ten fixed eight-step action branches are executed from a physics snapshot and compared with predictions from an eight-byte object-relative state. All arms share candidates and cost. The twelve held-out worlds were excluded from estimator training and calibration. Table~\ref{tab:tg2_unexposed} retains all worlds, including three whose repeated executions were not bit-identical (largest deviation 5.1 mm). Both pre-specified outcomes, pre-contact prediction and branch selection, favor the evolving state. A variant in which the object follows the tip after predicted re-contact lowers post-contact error but ranks branches less well, so sensory prediction and decision utility are reported separately.
\begin{table*}[t]\centering\small
\caption{\textbf{Tactile branch evaluation on twelve held-out worlds.} At each lost-contact anchor, ten fixed eight-step branches are executed and predicted from the same history. Top-1 is agreement with the best real branch; lower prediction error and regret are better. Contrasts use world-level bootstrap 95\% intervals. Two passes differ on three worlds (largest 5.1 mm); all worlds are retained.}
\label{tab:tg2_unexposed}
\begin{tabular}{lrrrr}\toprule
State update & Error before (mm) & Error after (mm) & Top-1 & Regret\\\midrule
Evolving, object held after contact & 2.44 & 4.36 & 0.64 & 0.0070\\
Evolving, object follows tip & 2.39 & 2.48 & 0.42 & 0.0201\\
Frozen state & 2.95 & 3.90 & 0.02 & 0.0658\\\bottomrule
\end{tabular}
\par\medskip
\begin{tabular}{llrc}\toprule
Comparison & Quantity & Difference [95\% CI] & Favorable worlds\\\midrule
Evolving $-$ frozen & Pre-contact error (mm) & $-0.51$ [$-0.73$, $-0.28$] & 10/12\\
Evolving $-$ frozen & Top-1 & $+0.62$ [$+0.43$, $+0.79$] & 11/12\\
Evolving $-$ frozen & Regret & $-0.0588$ [$-0.0698$, $-0.0467$] & 12/12\\
Follow $-$ object held & Post-contact error (mm) & $-1.87$ [$-2.06$, $-1.70$] & 12/12\\
Follow $-$ object held & Top-1 & $-0.21$ [$-0.30$, $-0.12$] & 1/12\\\bottomrule
\end{tabular}\end{table*}

\paragraph{Fixed-excursion recovery.}
Two fixed-excursion evaluations precede the varying-excursion protocol. The initial evaluation achieved 12.2 mm for evolved state, 6.1 mm for held and 4.8 mm for the feedback-law reference. Evolved minus held was 6.2 [2.3, 10.0] mm, positive on 8/12 worlds; the pre-specified criterion required 9/12. A forward-motion-conditioned write and a fitted outcome score were then selected using development worlds. Evaluation on twelve new worlds yielded 27.3 mm versus 3.4 mm for held and 0.2 mm for the law, a difference of 23.8 [21.4, 26.3] mm against held. A subsequent state-free command achieved 18.8 mm, giving an exploratory evolved-state difference of 8.5 [4.8, 12.7] mm. The varying-excursion protocol below includes stronger state-free references from the outset.

\paragraph{Varying-excursion evaluation.}
The subsequent study fits the chooser's outcome model and searches state-free commands on 36 development worlds, then evaluates 48 new worlds with sideways lengths 24, 48 and 72 steps. The primary reference receives excursion side; the stronger secondary command table receives side and length. Of 432 scanned conditions, 412 satisfy the validity and determinism checks, with at least one valid condition in every world. Differences are aggregated within world before computing paired $t$ intervals. Secondary intervals are unadjusted. The primary evolved--side-informed contrast is positive on 46/48 worlds. Against dead reckoning initialized at a constant pose, the difference is $-0.65$ [$-2.02$, $0.71$] mm. Against the private-pose chooser it is $+1.13$ [$-0.22$, $2.48$] mm; the private-pose chooser is a privileged reference, not an ordinary deployable baseline.

\begin{table*}[t]\centering\small
\caption{Real tactile frames (Real Tactile MNIST, GelSight Mini): late query after $T$ touches using an offline background computed from all 256 touches of the round; 48-byte payload plus up to 9 bytes of validity/age metadata; 100 held-out objects, mean $\pm$ std over 3 seeds (model selection on a fixed 50-round validation split). Q1 = digit identity, Q2 = material at a held-out position. Identity-query improvements over no history and proprioception are supported on all three seeds; material-query improvements are supported on two, with one unresolved. $B$ denotes the payload budget; six-byte slots occupy 12/48/126 bytes at budgets 16/48/128.}
\label{tab:real_tactile}
\resizebox{\textwidth}{!}{\begin{tabular}{lcccccc}\toprule
 & \multicolumn{3}{c}{Q1 accuracy} & \multicolumn{3}{c}{Q2 accuracy} \\ Arm & $T$=8 & $T$=16 & $T$=32 & $T$=8 & $T$=16 & $T$=32 \\ \midrule
byte-state memory (B=48) & 0.290 $\pm$ 0.010 & 0.354 $\pm$ 0.008 & 0.417 $\pm$ 0.020 & 0.656 $\pm$ 0.007 & 0.694 $\pm$ 0.004 & 0.700 $\pm$ 0.011 \\
128-dim. GRU (reference) & 0.356 $\pm$ 0.005 & 0.456 $\pm$ 0.016 & 0.485 $\pm$ 0.025 & 0.700 $\pm$ 0.003 & 0.719 $\pm$ 0.007 & 0.710 $\pm$ 0.008 \\
no history & 0.194 $\pm$ 0.009 & 0.192 $\pm$ 0.010 & 0.187 $\pm$ 0.010 & 0.667 $\pm$ 0.002 & 0.671 $\pm$ 0.002 & 0.659 $\pm$ 0.004 \\
proprioception only & 0.142 $\pm$ 0.005 & 0.152 $\pm$ 0.003 & 0.152 $\pm$ 0.006 & 0.636 $\pm$ 0.003 & 0.673 $\pm$ 0.001 & 0.683 $\pm$ 0.001 \\
\midrule
byte-state, B=16 & 0.271 $\pm$ 0.030 & 0.292 $\pm$ 0.025 & 0.305 $\pm$ 0.024 & \multicolumn{3}{c}{} \\
byte-state, B=48 & 0.290 $\pm$ 0.010 & 0.354 $\pm$ 0.008 & 0.417 $\pm$ 0.020 & \multicolumn{3}{c}{} \\
byte-state, B=128 & 0.311 $\pm$ 0.011 & 0.389 $\pm$ 0.019 & 0.452 $\pm$ 0.016 & \multicolumn{3}{c}{} \\
\bottomrule\end{tabular}}\end{table*}

\begin{table*}[t]\centering\small
\caption{\textbf{Synthetic pretraining for real tactile memory.} Differences in identity accuracy at 32 touches, in percentage points, over twenty initialization labels with paired 95\% $t$ intervals. Budgets are labeled real-training rounds; preprocessing uses within-window causal backgrounds.}
\label{tab:pretraining}
\setlength{\tabcolsep}{6pt}
\begin{tabular}{@{}rlrlcl@{}}\toprule
Budget & Contrast & \multicolumn{1}{c}{$\Delta$} & \multicolumn{1}{c}{95\% CI} & $n_{+}$ & Role \\
\midrule
25 & Synthetic pretraining $-$ none & $+2.06$ & [$0.70$, $3.43$] & 17/20 & Primary \\
25 & Synthetic pretraining $-$ Taxim & $+0.45$ & [$-0.74$, $1.64$] & 11/20 & Secondary \\
25 & Taxim $-$ none & $+1.61$ & [$0.01$, $3.22$] & 13/20 & Secondary \\
450 & Synthetic pretraining $-$ none & $+1.35$ & [$-0.54$, $3.24$] & 12/20 & Secondary \\
\bottomrule\end{tabular}
\end{table*}

\paragraph{Simulator and decision procedure.}
All arms select among a common stored set of candidate-path outcomes. Repeated PyBullet branches can vary with collision-pair ordering; the measured maximum deviation is 4.75 mm, typically below 0.4 mm, affecting a few percent of branches. The 412 accepted conditions pass the protocol's determinism check. Recovery executes the selected open-loop path and hands over to a feedback law at contact.

\paragraph{Real frames and preprocessing.}
The public Real Tactile MNIST dataset contains 600 printed objects with 256 touches each, collected with a GelSight Mini sensor on a robot manipulator. We retain its 500/100 train/test object split and use a fixed 50-round validation split for model selection. A median over all touches of a round supplies the offline background; the gate threshold is calibrated on training data. Consequently, ``after 32 touches'' describes the query sequence length, not a strictly causal preprocessing pipeline. Each slot stores two position bytes and four learned content bytes. A three-layer convolutional encoder with channels 16/32/64 and a 128-unit projection produces the code; normalized codes are quantized with a straight-through estimator and trained with the query heads. Writes merge within 4 mm, otherwise fill an empty slot or replace the oldest. The proprioception control keeps positions and the same image-derived write gate while zeroing the content bytes. The GRU reference uses a fixed 128-dimensional hidden state without quantization or the byte-memory budget. The full recoverable byte memory includes validity and age metadata. Table~\ref{tab:real_tactile} reports byte budgets, both query tasks and references. Illustrated frames are fixed-index samples from the first training object, selected independently of predictions or query success.

\section{Synthetic pretraining for real tactile memory}
\label{app:pretraining}
A supplementary experiment tests synthetic pretraining of the tactile encoder before real-frame learning, using twenty initialization labels and budgets of 25 and 450 labeled object rounds. It uses a causal background computed from frames observed within the query window, whereas the retention experiment in Sec.~\ref{sec:touch} uses an offline background. At budget 25, synthetic tactile pretraining increases 32-touch identity accuracy over no pretraining by 2.06 [0.70, 3.43] percentage points, positive on 17/20 labels. Its difference from Taxim-frame pretraining~\cite{si2022taxim} is 0.45 [$-0.74$, 1.64] points. At budget 450, its difference from no pretraining is 1.35 [$-0.54$, 3.24] points. Table~\ref{tab:pretraining} gives the primary and secondary comparisons.

\end{document}